\documentclass[journal]{IEEEtran}
\usepackage{amsmath,amsfonts}
\usepackage{amssymb,amsthm}
\usepackage{algorithm}
\usepackage{array}
\usepackage[caption=false,font=footnotesize,labelfont=sf,textfont=sf]{subfig} 

\usepackage{textcomp}
\usepackage{stfloats}
\usepackage{url}
\usepackage{verbatim}
\usepackage{graphicx}
\usepackage{cite}
\usepackage{bm}
\usepackage{color}
\usepackage{tabularx}
\usepackage{longtable}
\usepackage{tabu}
\usepackage[colorlinks,linkcolor=red]{hyperref}
\usepackage{multirow}
\newcolumntype{L}[1]{>{\raggedright\let\newline\\\arraybackslash\hspace{0pt}}m{#1}}
\newcolumntype{C}[1]{>{\centering\let\newline\\\arraybackslash\hspace{0pt}}m{#1}}
\newcolumntype{R}[1]{>{\raggedleft\let\newline\\\arraybackslash\hspace{0pt}}m{#1}}
\newcommand{\etal}{\textit{et al}.}
\newcommand{\ie}{\textit{i}.\textit{e}.}
\newcommand{\eg}{\textit{e}.\textit{g}.}

\usepackage{blkarray, bigstrut}
\usepackage{cleveref}
\usepackage{booktabs}

\definecolor{grey}{RGB}{130,130,130} 
\definecolor{black}{RGB}{0,0,0} 
\definecolor{red}{RGB}{0,0,0}
\usepackage{algpseudocode}
\DeclareMathOperator*{\argmax}{arg\,max}
\DeclareMathOperator*{\argmin}{arg\,min}

\usepackage{makecell}

\newcommand{\pair}[2]{\makecell[c]{#1 / #2}}

\makeatletter
\newcommand*{\rom}[1]{\expandafter\@slowromancap\romannumeral #1@}
\makeatother

\newenvironment{tightcenter}{%
  \setlength\topsep{3pt}
  \setlength\parskip{3pt}
  \begin{center}
  \begin{minipage}{.42\textwidth}
}{%
  \end{minipage}
  \end{center}
}

\begin{document}

\title{When No-Reference Image Quality Models Meet MAP Estimation in Diffusion Latents}

\author{Weixia~Zhang,~\IEEEmembership{Member,~IEEE},
        Dingquan~Li,~\IEEEmembership{Member,~IEEE},
        Guangtao~Zhai,~\IEEEmembership{Fellow,~IEEE},
        ~Xiaokang~Yang,~\IEEEmembership{Fellow,~IEEE},
        and~Kede~Ma,~\IEEEmembership{Senior~Member,~IEEE} 
\thanks{Weixia Zhang, Guangtao Zhai, and Xiaokang Yang are with the Institute of AI, School of Computer Science, Shanghai Jiao Tong University, Shanghai, China (e-mail: \{zwx8981, zhaiguangtao, xkyang\}@sjtu.edu.cn).}
\thanks{Dingquan Li is with the Department of Networked Intelligence, Pengcheng Laboratory, Shenzhen, China (e-mail: dingquanli@pku.edu.cn).}
\thanks{Kede Ma is with the Department of Computer Science and Shenzhen Research Institute, City University of Hong Kong, Kowloon, Hong Kong (e-mail: kede.ma@cityu.edu.hk).}
\thanks{Corresponding author: Kede Ma.}
}

\markboth{IEEE Transactions on Image Processing}%
{Zhang \MakeLowercase{\textit{et al.}}: When No-Reference Image Quality Models Meet MAP Estimation in Diffusion Latents}


\maketitle

\begin{abstract}
     Contemporary no-reference image quality assessment (NR-IQA) models can effectively quantify perceived image quality, often correlating strongly with human perceptual scores on standard IQA benchmarks. Yet few studies have explored their use as natural image priors for real-world image enhancement and compared them from a perceptual optimization perspective. In this work, we show, for the first time, that NR-IQA models can be plugged into the maximum a posteriori (MAP) framework for image enhancement. Specifically, we perform gradient-based optimization in the diffusion latent space rather than the raw pixel domain, leveraging a pretrained {differentiable} and {bijective} diffusion process. Different NR-IQA models yield different enhancement outputs, which facilitate model comparison from a new computational perspective. Guided by rankings obtained under the proposed framework, we develop an improved NR-IQA model by incorporating design elements from other highly ranked methods. Its enhancement outputs are preferred over those of its predecessors while maintaining high image fidelity.
\end{abstract}

\begin{IEEEkeywords}
No-reference image quality assessment, perceptual optimization, model comparison, diffusion models.
\end{IEEEkeywords}

\section{Introduction}
\IEEEPARstart{I}{mage} quality assessment (IQA) models have been extensively studied~\cite{wang2006modern} as computational proxies for human observers in evaluating perceived image quality. Depending on the availability of reference images, IQA models are typically classified into two categories: full-reference IQA (FR-IQA)~\cite{wang2004image, zhang2018unreasonable}  and no-reference IQA (NR-IQA)~\cite{mittal2012no, bosse2018deep}. FR-IQA models are most suitable when original, undistorted images are accessible for comparison.  However, in many real-world scenarios, identifying or obtaining ideal reference images is inherently difficult, if not impossible~\cite{cao2021debiased}. This limitation highlights the importance of NR-IQA approaches, which aim to assess perceived image quality in the absence of pristine references.

Beyond their role in performance evaluation, IQA models hold great potential for guiding the perceptual optimization of image processing systems.  FR-IQA models have long fulfilled this function, evolving from early formulations focused on error visibility and structural similarity~\cite{wang2004image} to more recent data-driven paradigms~\cite{zhang2018unreasonable,ding2022image}. A notable advance in this direction is the comprehensive comparison of FR-IQA models through the optimization of diverse image processing algorithms~\cite{ding2021comparison}, which effectively frames perceptual optimization within an analysis-by-synthesis framework~\cite{grenander2007pattern}. This naturally leads to the following question:
\begin{tightcenter}
    \textit{Can NR-IQA models be reliably employed for perceptual optimization, and should their performance be judged by comparing the quality of their respective optimized outputs?}
\end{tightcenter}

\begin{figure*}[!t]
    \centering
    \subfloat[\scriptsize $q_{\bm w}(\bm x) = 2.06$]{\includegraphics[width=0.24\textwidth]{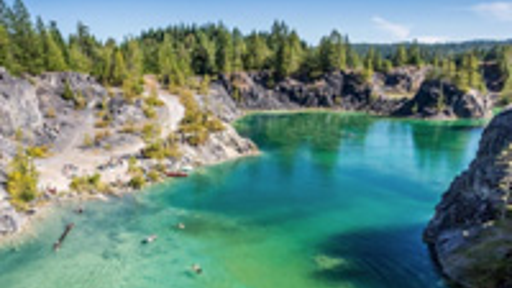}}\hskip.1em
    \subfloat[\scriptsize $q_{\bm w}(\bm x^\star) = 5.00$]{\includegraphics[width=0.24\textwidth]{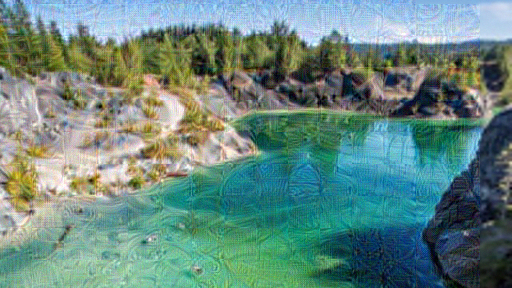}}\hskip.1em
    \subfloat[\scriptsize $q_{\bm w}(\bm x^\star) = 5.00$]{\includegraphics[width=0.24\textwidth]{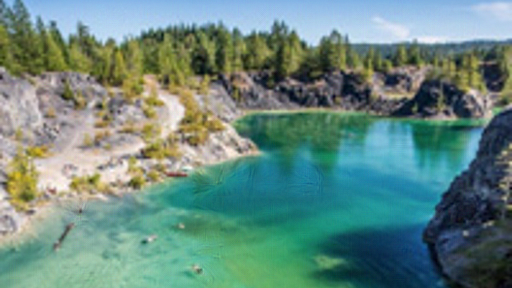}}\hskip.1em
    \subfloat[\scriptsize $q_{\bm w}(\bm x^\star) = 5.00$]{\includegraphics[width=0.24\textwidth]{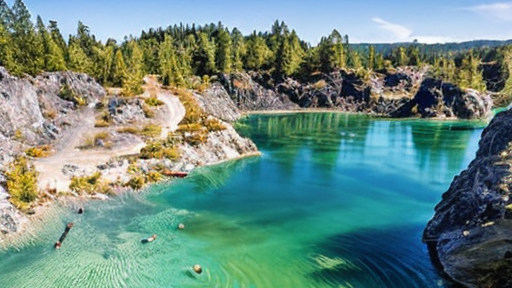}}
    \caption{\textbf{(a)} Distorted image used as the initial estimate. \textbf{(b)} Enhancement output obtained by directly maximizing a state-of-the-art NR-IQA model, LIQE~\cite{zhang2023blind} (see Eq.~\eqref{eq:eq1}). \textbf{(c)} Enhancement output obtained by MAP estimation (see Eq.~\eqref{eq:eq2}), where a reduced-resolution
Euclidean distance serves as the fidelity term and the LIQE score~\cite{zhang2023blind} defines the energy-based prior. \textbf{(d)} Enhancement output obtained by diffusion latent MAP estimation (see Eq.~\eqref{eq:mapdl}).  Each image is accompanied by its LIQE-predicted quality score, which ranges from zero to a maximum of five, with higher values indicating better predicted quality.}\label{fig:example}
\end{figure*}

A straightforward strategy for tackling this problem is to directly optimize an existing NR-IQA model $q_{\bm w}(\bm x):\mathbb{R}^M\mapsto\mathbb{R}$, parameterized by $\bm w$, with respect to the input image $\bm x$. Formally, one seeks
\begin{align}\label{eq:eq1}
    \bm x^\star = \argmax_{\bm x} q_{\bm w}(\bm x),
\end{align}
where larger values of $q_{\bm w}(\bm x)$ correspond to higher predicted quality, and $\bm x^\star \in \mathbb{R}^M$ denotes the optimized image. Because of the high dimensionality and non-convexity of this maximization problem, the fidelity of $\bm x^\star$ is implicitly constrained by the choice of initialization $\bm x^\mathrm{init}$. However, as noted in~\cite{zhang2022perceptual} and exemplified in Fig.~\ref{fig:example}(b), even when $q_{\bm w}(\cdot)$ is instantiated with a state-of-the-art NR-IQA model (\eg, LIQE~\cite{zhang2023blind}), the resulting optimized images often suffer from pronounced
local texture and color distortions, leading to perceptual degradation relative to the initial input. 

\begin{figure*}[!t]
    \centering
    \subfloat[$\lambda=1$]{\includegraphics[width=0.24\textwidth]{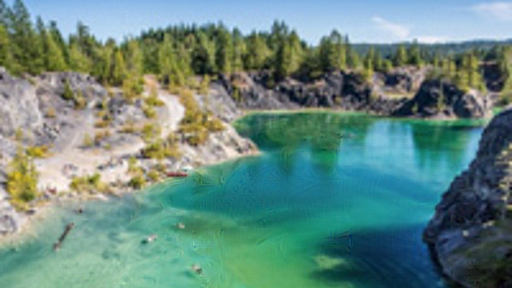}}\hskip.1em
    \subfloat[$\lambda=100$]{\includegraphics[width=0.24\textwidth]{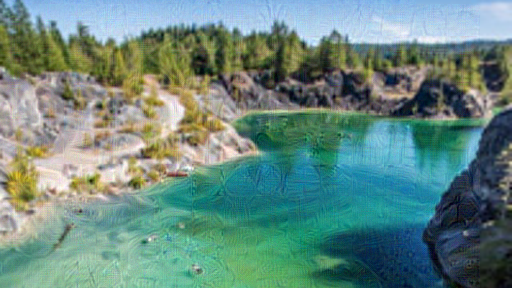}}\hskip.1em
    \subfloat[$\lambda=1,000$]{\includegraphics[width=0.24\textwidth]{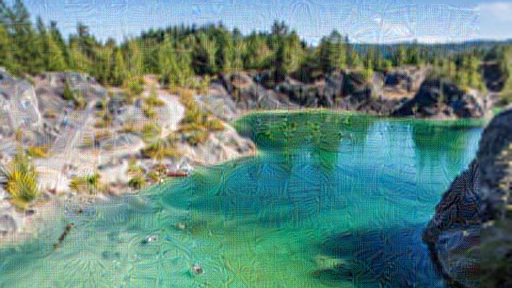}}\hskip.1em
    \subfloat[$\lambda=10,000$]{\includegraphics[width=0.24\textwidth]{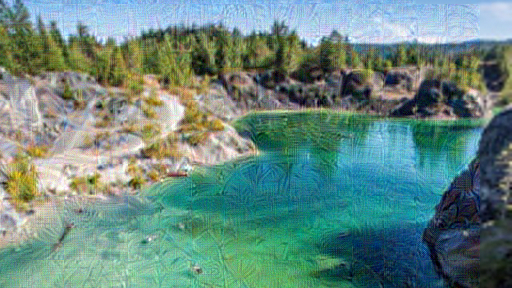}}
    \caption{Optimized images via MAP estimation (see Eq.~\eqref{eq:eq2}) for a range of $\lambda$ values.}\label{fig:adv_lambda}
\end{figure*}

A natural next step is to regard $q_{\bm w}(\cdot)$ as a prior over natural images and incorporate it into the standard maximum a posteriori (MAP) framework, yielding the optimization problem:
\begin{align}\label{eq:eq2}
    \bm x^\star &= \argmin_{\bm x} E(\bm x|\bm x^\mathrm{init}) \nonumber \\ &= \argmin_{\bm x}  D_0(\bm x, \bm x^\mathrm{init}) - \lambda q_{\bm w}(\bm x),
\end{align}
where the posterior $p(\bm x|\bm x^\mathrm{init}) \propto\exp\left(-E(\bm x|\bm x^\mathrm{init})\right)$ is expressed in terms of its corresponding Gibbs energy. Here, $D_0(\cdot,\cdot):\mathbb{R}^{M} \times \mathbb{R}^{M} \mapsto \mathbb{R}$ denotes the fidelity term (equivalently, the negative log-likelihood), which can be instantiated by an FR-IQA method. The hyperparameter $\lambda$ mediates the trade-off between the likelihood and prior. As shown in Fig.~\ref{fig:example}(c), even with an optimal choice of $\lambda$, the MAP-refined image, which corresponds to a visually imperceptible adversarial example of the NR-IQA model~\cite{zhang2022perceptual}, fails to surpass the quality of the initial estimate. As $\lambda$ increases, visual artifacts become increasingly pronounced (see Fig.~\ref{fig:adv_lambda}), eventually converging to the result obtained by directly maximizing $q_{\bm w}(\cdot)$.

In this work, we demonstrate that contemporary NR-IQA models can be employed for perceptual optimization within the MAP framework. Inspired by~\cite{wallace2023end}, we augment an NR-IQA method with a \textit{differentiable} and \textit{bijective} diffusion transform. This integration endows the model with generative capacity, enabling it to capture the rich diversity of natural images through operations in the diffusion latent space\footnote{Here, we distinguish between ``latent diffusion,'' where the diffusion process occurs in a latent space rather than the raw pixel domain, and ``diffusion latent,'' which treats the diffusion process as a feature transform, with the pure noise vector serving as the latent representation.}. Specifically, we adopt the method of exact diffusion inversion via coupled transformations (EDICT)~\cite{wallace2023edict}, which leverages affine coupling~\cite{DinhSB17} to establish a bijection between the diffusion model's input latent and its corresponding noise latent.

By performing MAP estimation in the diffusion latent space, we arrive at a new computational approach for comparing NR-IQA models within an analysis-by-synthesis framework. In particular, we embed eight representative NR-IQA models into this framework to produce enhancement outputs for a set of photographic images degraded by real-world visual artifacts~\cite{ghadiyaram2016massive, hosu2020koniq, fang2020perceptual}. Using \textit{debiased} psychophysical testing~\cite{cao2021debiased} on these outputs, we assess the relative optimization capabilities of these quality models in this controlled setting. The results offer complementary insights for NR-IQA model development beyond conventional fixed-set, correlation-based evaluations.

Guided by our psychophysical experiment, we improve LIQE~\cite{zhang2023blind}---a leading NR-IQA model in this setting---by incorporating the multi-scale image representation strategy of another highly ranked model, MUSIQ~\cite{ke2021musiq}, and retraining it on a collection of IQA datasets~\cite{sheikh2006statistical, larson2010most, ciancio2011no, ghadiyaram2016massive, hosu2020koniq, lin2019kadid}. The resulting LIQE variant exhibits improved optimization capability under the proposed framework on real-world images affected by complex distortions.


In summary, our contributions are threefold.
\begin{itemize}
    \item  We propose diffusion latent MAP estimation, enabling existing NR-IQA models to serve as energy-based image priors for perceptual optimization in real-world image enhancement.
    \item Within this framework, we conduct a systematic comparison of eight contemporary NR-IQA models, leading to several key insights.
    \item Building on these insights, we improve a leading NR-IQA model  by combining design elements from multiple \textcolor{red}{highly ranked models within the proposed framework}, achieving stronger optimization performance.
\end{itemize}

\begin{figure*}[!t]
    \centering
        \subfloat[]{\includegraphics[width=0.48\textwidth]{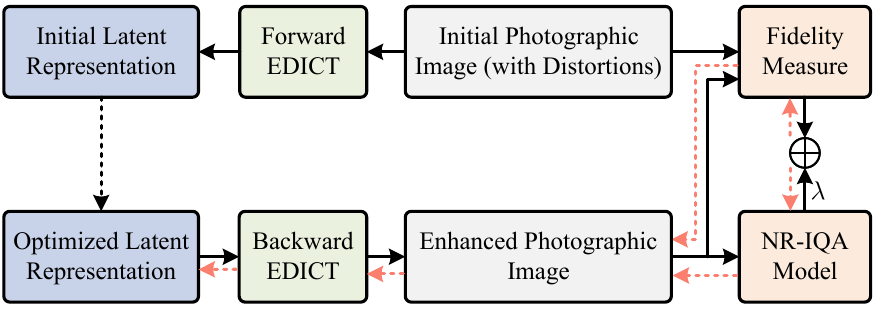}}\hskip.5em
      \subfloat[]{\includegraphics[width=0.48\textwidth]{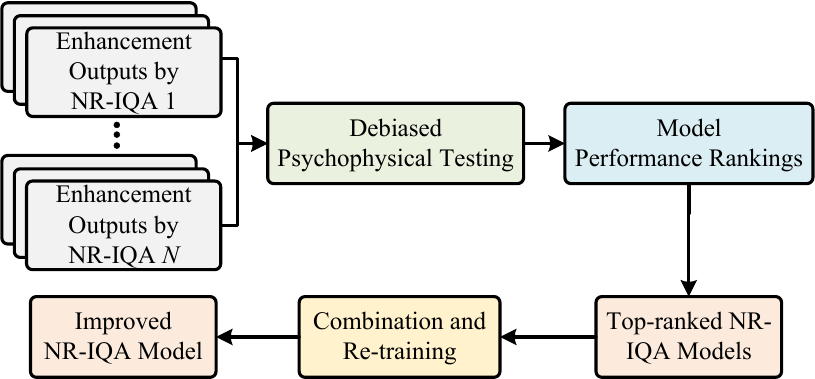}}
    \caption{\textbf{(a)} System diagram of diffusion latent MAP estimation. \textbf{(b)} NR-IQA model comparison and improvement enabled by the proposed diffusion latent MAP estimation framework. }\label{fig:framework}
\end{figure*}

\section{Related Work}
\label{sec:related}
In this section, we first review recent advances in NR-IQA. Next, we discuss approaches for model comparison through the lens of analysis-by-synthesis~\cite{mumford1994pattern, grenander2007pattern}. Finally, we provide an overview of diffusion models, highlighting their applications across various computer vision tasks.

\subsection{NR-IQA Models}
\label{sec:nriqa}
Early NR-IQA methods predominantly combined handcrafted natural scene
statistics~\cite{mittal2012no, mittal2013making,
ghadiyaram2017perceptual}
with regression models to predict perceptual quality. While effective under constrained conditions, these methods struggled to generalize across the diverse image content and distortion types encountered in the real world. The advent of deep learning shifted the paradigm toward end-to-end training of deep neural networks (DNNs), reducing reliance on manually
designed features. Nevertheless, the performance of these models remains constrained by the limited availability of human perceptual data, often represented as mean opinion scores (MOSs). To mitigate this challenge, various data-efficient strategies have been proposed, including quality-aware pretraining~\cite{Ma2018End, zhang2020blind}, pseudo-label learning~\cite{ma2019blind, wu2020end}, meta-learning~\cite{zhu2020metaiqa}, contrastive learning~\cite{madhusudana2022image, agnolucci2024arniqa}, and multitask learning~\cite{fang2020perceptual, zhang2023blind}. Building on these efforts, recent research has increasingly focused on more practical and robust solutions, such as predicting spatially varying quality~\cite{ying2020from}, adapting to subpopulation shifts~\cite{zhang2021uncertainty, zhang2023continual, zhang2024task, li2024bridging}, designing more powerful network architectures~\cite{ke2021musiq, yang2022maniqa, chen2024topiq}, and exploiting multimodal information~\cite{wang2023exploring, wu2024q, wu2024comprehensive}.

\subsection{Model Comparison Methods}
\label{sec:analysis_by_synthesis}
Model comparison is a central component of model development~\cite{anderson2004model}. The conventional approach involves constructing a static test set with ground-truth annotations and assessing how well model predictions align with these annotations. Beyond this discriminative paradigm, models can also be tested in a generative manner, as exemplified by the analysis-by-synthesis framework in Grenander's Pattern Theory~\cite{mumford1994pattern, grenander2007pattern}. Representative generative model comparison methods include the maximum differentiation (MAD) competition~\cite{wang2008maximum}, eigen-distortion analysis~\cite{berardino2017eigen},  and controversial stimuli synthesis~\cite{golan2020controversial}. Conceptually, the synthesis process can also be reformulated as discrete sample selection from large-scale unlabeled image collections, extending its applicability to a broad range of tasks, such as IQA~\cite{ma2020group}, image recognition~\cite{wang2020going}, semantic segmentation~\cite{yan2021exposing}, image enhancement~\cite{cao2021debiased}, and large-scale language modeling~\cite{feng2024sample}. 

Our work is closely related to perceptual attacks on NR-IQA
models~\cite{zhang2022perceptual}, as both can be framed as MAP
estimation.
However, they differ substantially in goal and implementation:
perceptual attacks evaluate adversarial robustness to imperceptible
pixel-level perturbations, whereas our method operates in the diffusion
latent space to assess how effectively NR-IQA models can guide
real-world image enhancement.
\textcolor{red}{In other words, perceptual attacks seek changes in
quality predictions under perceptual invariance, whereas our framework
compares the perceptual quality of outputs generated by different
model-guided optimization objectives.}

\subsection{Diffusion Models}
\label{sec:dms}
Diffusion models constitute a prominent class of generative models designed to reverse a gradual noise-injection process that unfolds over multiple steps~\cite{croitoru2023diffusion}. The literature broadly distinguishes three major formulations: 1) denoising diffusion probabilistic models~\cite{sohl2015deep, ho2020denoising}, inspired by principles of non-equilibrium thermodynamics~\cite{pokrovskii2020thermodynamics}; 2) noise-conditioned score networks~\cite{song2019generative}, trained via score matching~\cite{hyvarinen2005estimation} to estimate the gradients of the log-likelihood of noisy data; and 3) stochastic differential equations~\cite{song2021score}, which offer a unifying perspective that generalizes the first two formulations. 

To enhance controllability, classifier guidance~\cite{dhariwal2021diffusion} and classifier-free guidance~\cite{ho2021classifier} have been proposed, though they require training noise-aware classifiers and class-conditional diffusion models, respectively.

In photographic image editing~\cite{HertzMTAPC23}, \textit{diffusion inversion}---the recovery of the noise vector that produces a target image via reverse diffusion---is a critical prerequisite. 
EDICT~\cite{wallace2023edict} provides an elegant solution by using
affine coupling~\cite{DinhSB17} to construct a differentiable bijection
between a diffusion model's input latent and its corresponding noise
latent. In contrast to alternatives such as denoising diffusion implicit models~\cite{SongME21}, EDICT supports exact diffusion inversion without additional training or knowledge distillation. Building on this insight~\cite{wallace2023end}, we integrate NR-IQA models with EDICT to enable MAP estimation in the diffusion latent space for real-world image enhancement. 

Existing diffusion-based image enhancers typically augment pretrained diffusion models with auxiliary modules~\cite{wang2024exploiting, lin2024diffbir, yu2024scaling, wu2024seesr}, modify the sampling procedure~\cite{kawar2022denoising, wang2023zero, chung2023diffusion, zhu2023denoising}, or combine both strategies. \textcolor{red}{Recent IQA-guided approaches follow a different route, employing quality-adaptive refinement~\cite{li2024q}, training quality-conditioned diffusion adapters~\cite{abud2025iqa}, or using NR-IQA objectives to guide task-specific training~\cite{zhang2025augmenting}. In contrast to both lines of work, our framework directly optimizes each input in the diffusion latent space using existing NR-IQA and diffusion models, without additional training or fine-tuning. Through EDICT, it is also compatible with modified diffusion backbones and can be deployed as a post-enhancement step to refine the outputs of existing enhancers.}

\section{Diffusion Latent MAP Estimation}
\label{sec:method}
In this section, we first present the necessary preliminaries of diffusion models, with particular emphasis on EDICT~\cite{wallace2023edict}. We then introduce the proposed diffusion latent MAP estimation framework for real-world image enhancement, and employ it to compare NR-IQA models. Fig.~\ref{fig:framework} shows the system diagram.

\subsection{Preliminaries}
\label{subsec:pre}
For a photographic image $\bm{x}\in\mathbb{R}^M$, potentially subject to complex, real-world distortions, the objective of an NR-IQA model $q_{\bm w}(\bm x)$ is to estimate the perceived quality of $\bm{x}$ in a manner that aligns closely with its underlying MOS $q(\bm{x})\in\mathbb{R}$. 

To manipulate images, we also work with a diffusion model~\cite{yang2023diffusion}. In the forward diffusion process, a sequence of noise-schedule coefficients $\{\alpha_{t}\}_{t=0}^{T}$ is defined, where $\alpha_{0} = 1$, $\alpha_{T} \approx 0$, and $\alpha_t$ decreases monotonically with $t$. At the $t$-th step, the input image $\bm x_0$ is progressively corrupted as 
\begin{align}\label{eq:noising}
\bm x_{t} = \sqrt{\alpha_{t}}\bm x_0 + \sqrt{1-\alpha_{t}}\bm \epsilon,
\end{align}
where $\bm\epsilon \sim \mathcal{N}(\bm{0}, \mathbf{I})$. The reverse diffusion process corresponds to data generation, where a denoising network $\bm \epsilon_{\bm\theta}(\cdot;t):\mathbb{R}^M\mapsto\mathbb{R}^M$, parameterized by $\bm \theta$ and conditioned on $t$, is trained to progressively remove the injected noise~\cite{SongME21}:
\begin{align}\label{eq:pd}
    \hat{\bm x}_{t-1} =&  \sqrt{\alpha_{t-1}} \frac{\hat{\bm{x}}_t -\sqrt{1-\alpha_{t}}\bm\epsilon_{\bm\theta}(\hat{\bm x}_t;t)}{\sqrt{\alpha_t}} + \sqrt{1-\alpha_{t-1}}\bm \epsilon_{\bm\theta}(\hat{\bm x}_t;t)\nonumber\\
    =& a_t \hat{\bm x}_t + b_t\bm\epsilon_{\bm\theta}(\hat{\bm x}_t;t), 
\end{align}
where $a_{t} = \sqrt{{\alpha_{t-1}}/{\alpha_{t}}}$, $b_{t} = -\sqrt{{\alpha_{t-1}(1-\alpha_{t})}/{\alpha_{t}}} + \sqrt{1-\alpha_{t-1}}$, and $\hat{\bm x}_T = \bm \epsilon$. The effectiveness of diffusion models largely relies on the denoising network $\bm \epsilon_{\bm\theta}(\cdot;t)$\textcolor{black}{~\cite{ho2020denoising, nichol2021improved, croitoru2023diffusion, chang2026design}}, which is optimized by minimizing
\begin{equation}\label{eq:ldm}
    \mathbb{E}_{\bm x_0, \bm\epsilon \sim \mathcal{N}(\bm{0}, \mathbf{I}),t} \left[ \lambda(t)\lVert \bm \epsilon - \bm \epsilon_{\bm \theta}(\bm x_{t}; t)\rVert_{2}^{2} \right],
\end{equation}
where $\lambda(t)$ denotes a positive time-dependent weighting factor, and $\bm x_t$ is generated via Eq.~\eqref{eq:noising}. However, this denoising process is not strictly invertible because it relies on the linearization assumption $\bm \epsilon_{\bm \theta}(\hat{\bm x}_{t}; t) = \bm \epsilon_{\bm \theta}(\hat{\bm x}_{t-1}; t)$, which does not hold exactly. 

Recognizing the resemblance between diffusion-based denoising and affine coupling in normalizing flows~\cite{DinhSB17}, Wallace~\etal~\cite{wallace2023edict} achieved invertibility in~Eq.~\eqref{eq:pd} by introducing two coupled noise vectors (and their intermediate states):
\begin{align}
    {\bm x}_{t-1} &= a_t{\bm x}_t + b_t\bm \epsilon_{\bm \theta}({\bm y}_t; t),\label{eq:ni1}\\
     {\bm y}_{t-1} &= a_t{\bm y}_t + b_t\bm \epsilon_{\bm \theta}({\bm x}_{t-1}; t),\,\, t\in\{1,\ldots, T\},\label{eq:ni2}
\end{align}
with initialization ${\bm x}_T = {\bm y}_T =\bm \epsilon$. Here, we omit the ``hat'' symbol to emphasize the bijective nature of Eqs.~\eqref{eq:ni1} and~\eqref{eq:ni2}. To improve numerical stability and preserve fidelity to the original diffusion dynamics, a mixing operation is further introduced, leading to the denoising updates:
\begin{align}
{\bm{x}}_{t}' &= a_{t} {\bm x}_{t} + b_{t} \bm \epsilon_{\bm \theta}({\bm y}_{t}; t),\label{eq:ii1}  \\
{\bm y}_{t}' &= a_{t}{\bm y}_{t} + b_{t} \bm \epsilon_{\bm \theta}({\bm x}_{t}'; t),  \\
{\bm x}_{t-1} &= \eta {\bm x}_{t}' + (1-\eta) {\bm y}_{t}', \label{eq:ii3} \\
{\bm y}_{t-1} &= \eta {
\bm y}_{t}' + (1-\eta) {\bm x}_{t-1}\label{eq:ii4},\,\, t\in\{1,\ldots, T\},
\end{align}
where $\eta \in [0, 1]$ is a mixing parameter that mitigates deviations from the linearization assumption. 

Linearly inverting Eqs.~\eqref{eq:ii1} to~\eqref{eq:ii4} defines a deterministic noising process that acts as a feature transform:
\begin{align}\label{eq:edict_inverse}
{\bm y}_{t}' &= ({\bm y}_{t-1} - (1-\eta)  {\bm x}_{t-1} )/\eta , \\
{\bm x}_{t}' &= ({\bm x}_{t-1} - (1-\eta) {\bm y}_{t}') /\eta , \\
{\bm y}_{t} &= ({\bm y}_{t}' - b_{t} \bm \epsilon_{\bm \theta}({\bm x}_{t}'; t)) / a_{t}, \\
{\bm x}_{t} &= ({\bm x}_{t}' - b_{t}\bm\epsilon_{\bm \theta}({\bm y}_{t}; t)) / a_{t},\,\, t\in\{1,\ldots, T\}.
\end{align}
We collectively denote the resulting differentiable and bijective mapping from the coupled input representations $(\bm x_0,\bm y_0)$, with $\bm y_0=\bm x_0$, to their noise representations $(\bm x_T,\bm y_T)$ by $\bm d(\cdot):\mathbb{R}^{2M}\mapsto\mathbb{R}^{2M}$. Its inverse is denoted by $\bm h(\cdot)=\bm d^{-1}(\cdot)$. When a latent diffusion model is used, $\bm x_0$ and $\bm y_0$ represent variational autoencoder (VAE) latents, and the VAE encoder and decoder lie outside EDICT's bijectivity guarantee.

 \subsection{Diffusion Latent MAP Estimation}
\label{subsec:map}
In Bayesian statistics, the MAP estimate of an unknown quantity is defined as the mode of its posterior distribution. In the context of image enhancement, MAP estimation can be formulated as 
\begin{align}\label{eq:perceptual_attack}
\bm{x}^\star &=\mathop{\arg\max}_{\bm x}  p(\bm{x}|\bm{x}^\mathrm{init})=\mathop{\arg\max}_{\bm x}  p(\bm x^\mathrm{init}|\bm{x})p(\bm x)  \nonumber \\
&= \mathop{\arg\min}_{\bm x}  -\log{p(\bm{x}^\mathrm{init}|\bm{x})} - \log{p(\bm{x})}.
\end{align}
The first term measures the fidelity of the \textcolor{red}{enhancement output} $\bm x$ relative to the initial input $\bm x^\mathrm{init}$. A common choice is the mean squared error (MSE), which is differentiable and straightforward to optimize.
The second term enforces naturalness by incorporating image priors or regularizers. Typical examples include total variation regularization~\cite{rudin1992nonlinear},  Gaussian scale mixture priors~\cite{portilla2003image}, (non-local) patch-based priors~\cite{buades2005non},  sparsity regularization~\cite{wright2010sparse},  low-rank priors~\cite{cai2010singular}, deep image priors~\cite{ulyanov2020deep}, and adversarial losses~\cite{goodfellow2014generative}.

 \begin{algorithm}[!t]
\caption{Diffusion Latent MAP Estimation Solver}
\label{algoe}
{\bf Require:}
An NR-IQA model $q_{\bm w}(\cdot)$, an FR-IQA model defining the image fidelity measure $D_0(\cdot,\cdot)$, an EDICT model to map a photographic image into a diffusion latent space $\bm{d}(\cdot)$ and its inversion $\bm{h}(\cdot)$, the number of optimization steps $\mathrm{MaxIter}$, the number of diffusion steps $T$, the trade-off parameter $\lambda$,  the momentum parameter $\rho$, and the learning rate $\alpha_{\mathrm{lr}}$\\
{\bf Input:}
An initial image $\bm{x}^\mathrm{init}$\\
{\bf Output:} 
A pair of enhancement outputs $[\bm{x}^{\star}, \bm{y}^\star$] with nearly identical content and appearance 
\begin{algorithmic}[1]
    \State $\bm m_{x} = \bm {0}, \bm{m}_{y}=\bm {0}$
    \State $[{\bm x}_T, {\bm y}_{T}] \leftarrow \bm{d}(\bm x^\mathrm{init}, \bm x^\mathrm{init}) $ 
    \For {$i = 1\to \mathrm{MaxIter}$}
        \State $[{\bm x}, {\bm y}] \leftarrow \bm{h}({\bm x}_T, {\bm y}_T) $ 
        \State $\ell(\bm x_T) \leftarrow \ D_0({\bm x}, \bm{x}^\mathrm{init}) - \lambda q_{\bm w}({\bm x})$ 
        \State $\ell(\bm y_T) \leftarrow \ D_0({\bm y}, \bm{x}^\mathrm{init}) - \lambda q_{\bm w}({\bm y})$
        \State $\ell(\bm x_T, \bm y_T) \leftarrow (\ell(\bm x_T) + \ell(\bm y_T))/2$
        \State $\Delta \ell (\bm x_T) \leftarrow\partial \ell(\bm x_T, \bm y_T) /\partial \bm x_T$  
        \State $\Delta \ell (\bm y_T) \leftarrow\partial \ell(\bm x_T, \bm y_T) /\partial \bm y_T$ 
        \State $[\bm m_x, \bm m_y] \leftarrow \rho [\bm m_x, \bm m_y] - \alpha_\mathrm{lr}[\Delta \ell(\bm x_T), \Delta \ell(\bm y_T)]$ 
        \State $[\bm{x}_{T}, \bm{y}_{T}] \leftarrow [\bm{x}_{T}, \bm{y}_{T}] + [\bm m_x, \bm m_y]$
    \EndFor
    \State $[\bm{x}^{\star},\bm{y}^\star] \leftarrow \bm{h}(\bm{x}_{T}, \bm{y}_T)$
\end{algorithmic}
\end{algorithm}

As discussed in the Introduction, NR-IQA scores can be incorporated into MAP estimation as image priors, favoring outputs with higher predicted quality while the fidelity term constrains deviations from the initial estimate $\bm x^\mathrm{init}$. However, conventional NR-IQA models are typically trained in a \textit{discriminative} manner on finite datasets with restricted distortion types, which limits their reliability as \textcolor{red}{standalone} naturalness priors (see Fig.~\ref{fig:example}). To address this limitation, inspired by~\cite{wallace2023end}, we combine each NR-IQA model with the differentiable and bijective diffusion transform $\bm h(\cdot)$ described in Sec.~\ref{subsec:pre}. This combination introduces complementary generative structure and enables MAP estimation in the diffusion latent space:  
\begin{align}\label{eq:mapdl}
\bm{x}_T^{\star}
=
\mathop{\arg\min}_{\bm{x}_T}
D_{0}\left(\bm h(\bm{x}_T), \bm{x}^\mathrm{init}\right)
-
\lambda q_{\bm w}(\bm h(\bm{x}_T)),
\end{align}
where we omit the coupled noise vector $\bm y_T$ for notational simplicity. \textcolor{red}{This objective admits a direct energy-based interpretation. Specifically, the fidelity term defines the likelihood
$p(\bm x^\mathrm{init}|\bm h(\bm{x}_T))\propto
\exp\!\left[-D_0(\bm h(\bm{x}_T),\bm x^\mathrm{init})\right]$,
whereas the NR-IQA score defines the unnormalized prior
$p(\bm h(\bm{x}_T))\propto\exp\!\left[\lambda q_{\bm w}(\bm h(\bm{x}_T))\right]$.
Consequently, the objective in Eq.~\eqref{eq:mapdl} corresponds, up to an additive constant, to a negative log-posterior and assigns greater prior density to images with higher predicted quality.}
We solve Eq.~\eqref{eq:mapdl} using the gradient-based procedure summarized in Algorithm~\ref{algoe}. The final outputs,
$[\bm x^\star,\bm y^\star]
=\bm h(\bm x_T^\star,\bm y_T^\star)$,
are nearly identical in content and appearance, consistent with the observations in~\cite{wallace2023end}. \textcolor{red}{We use $\bm x^\star$ as the enhancement output.}

\begin{figure*}[t]
    \centering
   {\includegraphics[width=0.95\textwidth]{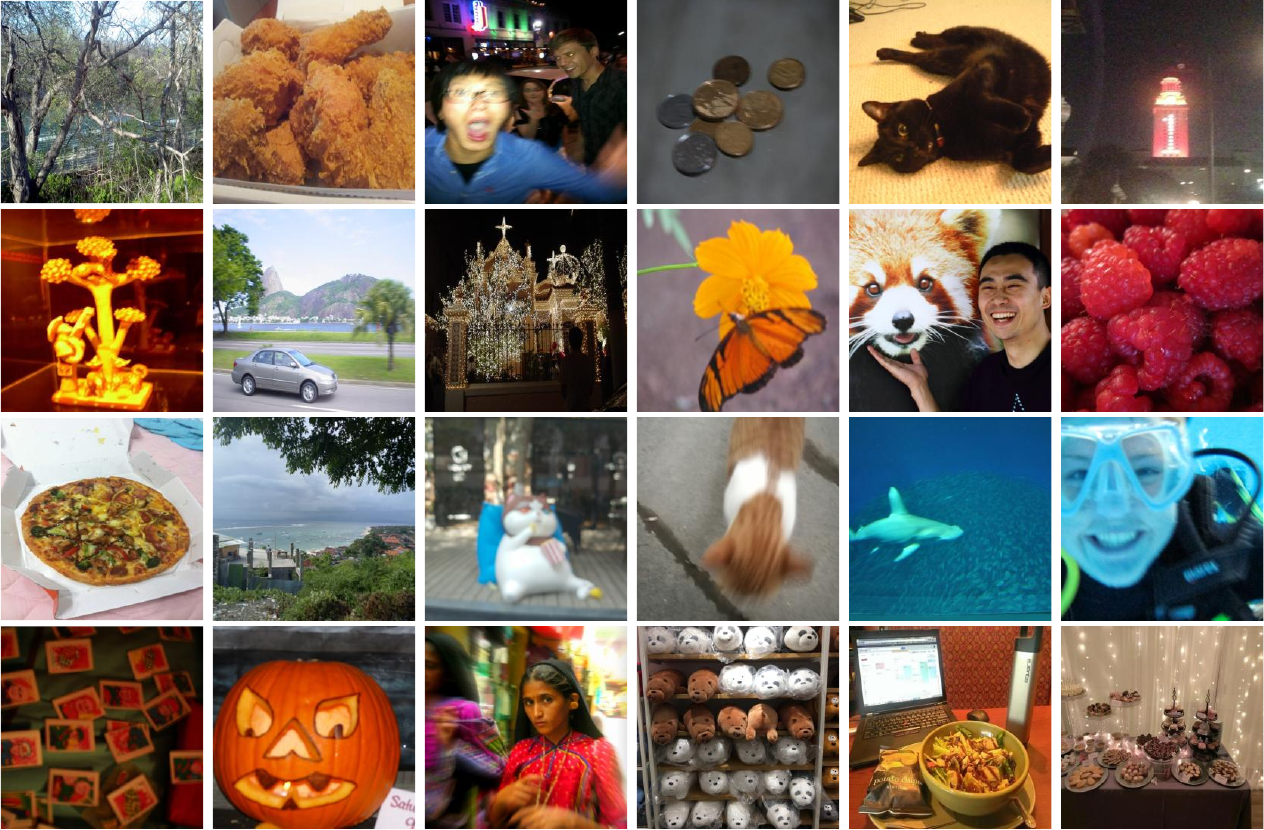}}
    \caption{Representative photographic images used as inputs to the proposed enhancement framework.}\label{fig:ref_imgs}
\end{figure*}

\subsection{NR-IQA Model Comparison}
\label{subsec:subjective}
By incorporating different NR-IQA models into diffusion latent MAP estimation (Eq.~\eqref{eq:mapdl}), we obtain multiple enhancement outputs for the same input image, with distinct characteristics in color and detail reproduction. This variation motivates a principled comparison of NR-IQA models within the proposed framework. Specifically, we compile a candidate set of photographic images with complex distortions, denoted by $\mathcal{X}=\{\bm x^{(i)}\}_{i=1}^{\vert\mathcal{X}\vert}$. For each of the $N$ NR-IQA models $\{q_j(\cdot)\}_{j=1}^{N}$, we apply diffusion latent MAP estimation to every image in $\mathcal{X}$, collectively producing $N\vert\mathcal{X}\vert$ enhancement outputs, $\{\{\bm r_j(\bm x^{(i)})\}_{i=1}^{\vert\mathcal{X}\vert}\}_{j=1}^{N}$. Here, $\bm r_j(\cdot):\mathbb{R}^M\mapsto\mathbb{R}^M$ denotes the complete enhancement mapping obtained by solving Eq.~\eqref{eq:mapdl} using the $j$-th NR-IQA model.

Exhaustive psychophysical evaluation of all outputs would be prohibitively costly. We therefore adopt the efficient and debiased testing procedure in~\cite{cao2021debiased}. For each pair of NR-IQA models $q_i(\cdot)$ and $q_j(\cdot)$, with corresponding enhancement mappings $\bm r_i(\cdot)$ and $\bm r_j(\cdot)$, we select a compact subset of $K$ input images, where $K\ll\vert\mathcal{X}\vert$. We first select $\tilde{\bm x}^{(1)}$ by maximizing $D_1(\bm r_i(\bm x),\bm r_j(\bm x))$ over $\bm x\in\mathcal{X}$, where $D_{1}(\cdot,\cdot):\mathbb{R}^{M}\times \mathbb{R}^{M} \mapsto \mathbb{R}$ measures the perceptual distance between two enhancement
outputs, and initialize $\mathcal{S}^{(1)}=\{\tilde{\bm x}^{(1)}\}$. For $k\geq1$, the next image is selected by solving
\begin{align} \label{eq:mad}
    \tilde{\bm x}^{(k+1)} &= \argmax_{\bm x \in \mathcal{X}\setminus\mathcal{S}^{(k)}} D_{1}\left(\bm r_{i}(\bm x), \bm r_{j}(\bm x)\right) + \gamma D_{2}(\bm{x}, \mathcal{S}^{(k)}),
\end{align}
where $\mathcal{S}^{(k)}=\{\tilde{\bm x}^{(k')}\}_{k'=1}^{k}$ denotes the set of $k$ previously selected images.
 $D_{2}(\cdot,\cdot):\mathbb{R}^M\times\bigl(\mathcal{P}(\mathbb{R}^M)\setminus\{\emptyset\}\bigr)\mapsto\mathbb{R}$, where $\mathcal{P}(\cdot)$ is the power set function, quantifies the semantic distance between an image and a nonempty set of images, thereby promoting content diversity. The hyperparameter $\gamma$ balances these two terms. 
 \textcolor{red}{Following~\cite{cao2021debiased}, we compute $D_1(\cdot,\cdot)$ using DISTS~\cite{ding2022image} and define $D_2(\cdot,\cdot)$ as the mean pairwise MSE between the last-layer VGGNet~\cite{simonyan2014very} feature representations of the candidate image $\bm x$ and images in the selected set $\mathcal S^{(k)}$.} Once $\tilde{\bm x}^{(k+1)}$ is identified, it is added to $\mathcal{S}^{(k)}$  to form  $\mathcal{S}^{(k+1)}$.

Consequently, we generate $KN(N-1)/2$ image pairs for formal psychophysical testing using a two-alternative forced choice (2AFC) paradigm. In each trial, subjects are shown two \textcolor{red}{enhancement outputs} with identical content, each produced using a different NR-IQA model, and indicate which output they perceive to be of higher quality. Further details of the experimental protocol are provided in Sec.~\ref{subsec:selection}.

We aggregate the human preference judgments into an $N\times N$ \textcolor{red}{count matrix} $\mathbf C$, where $C_{ij}$ denotes the number of valid judgments favoring the output guided by $q_i(\cdot)$ over that guided by $q_j(\cdot)$. Under the Thurstone Case V model~\cite{tsukida2011analyze}, we represent the latent quality scores of the $N$ models by $\bm\mu=[\mu_1,\mu_2,\ldots,\mu_N]^{\intercal}$ and estimate them by constrained maximum likelihood:
\begin{align}\label{eq:jod}
{\bm\mu}^\star
&=
\underset{\bm\mu:\,\sum_{i=1}^{N}\mu_i=0}{\arg\max}\;
\ell(\bm\mu;\mathbf C), \nonumber\\
\ell(\bm\mu;\mathbf C)
&=
\sum_{\substack{i,j=1\\i\neq j}}^{N}
C_{ij}\log\Phi\left(
\frac{\mu_i-\mu_j}{\sqrt{2}\sigma}
\right),
\end{align}
where $\Phi(\cdot)$ denotes the standard Gaussian cumulative distribution function. The zero-sum constraint resolves the translation ambiguity inherent in the relative scores. Following~\cite{perez2020pairwise}, we fix the observer-model standard deviation at $\sigma=1.0484$, such that a one-unit difference on the just-objectionable-difference (JOD) scale corresponds to a $75\%$ probability that one output is preferred over the other. The global ranking is obtained by ordering the estimated scores ${\bm\mu}^\star$.

\section{NR-IQA Model Improvement}\label{subsec:finetune}
The proposed comparison framework not only ranks NR-IQA models but also reveals the design choices associated with strong
enhancement performance. We therefore use these findings to guide
the development of an improved NR-IQA model. As will be clear in Sec.~\ref{subsec:results}, LIQE~\cite{zhang2023blind}, MUSIQ~\cite{ke2021musiq}, and UNIQUE~\cite{zhang2021uncertainty}
consistently outperform the other models within the proposed
framework. Building on their complementary strengths, our model
incorporates 1) the pretrained CLIP backbone of LIQE,
2) the multi-scale image representations of MUSIQ, and
3) the mixed-dataset training strategy adopted by LIQE and UNIQUE.

\subsection{Improved Model Design}
The original LIQE associates an input image $\bm x$ to multiple textual templates and predicts joint distributions for three tasks: NR-IQA, scene classification, and distortion type identification. Here, we focus exclusively on NR-IQA and represent image quality using the five-point Likert scale $\mathcal{V}=\{1,2,3,4,5\}$, whose levels correspond to ``bad,'' ``poor,'' ``fair,'' ``good,'' and ``perfect,''  respectively. The discrete  quality distribution is converted into the continuous
quality estimate:
\begin{align}\label{eq:stl}
    q_{\bm w}(\bm{x}) = \sum_{v=1}^{\vert\mathcal{V}\vert}{p}_{\bm w}(v|\bm{x})\times v,
\end{align}
where ${p}_{\bm w}(v|\bm x)$ is the predicted probability that image $\bm x$ belongs to quality level $v$. Built on the CLIP backbone~\cite{radford2021learning}, LIQE comprises an image encoder $\bm{f}_{\bm{\phi}}(\cdot):\mathbb{R}^M\mapsto\mathbb{R}^L$  and a language encoder $\bm{g}_{\bm{\varphi}}(\cdot): \mathcal{T} \mapsto \mathbb{R}^L$, which map an image crop and a textual prompt, respectively, to
$L$-dimensional embeddings. Here, $\mathcal T$ denotes the set of
textual prompts, and the model parameters are collected as
$\bm w=(\bm\phi,\bm\varphi)$.

The image encoder extracts a visual embedding matrix
$\bm F(\bm x)\in\mathbb R^{U\times L}$
from $U$ sub-images of size $224\times224\times3$ cropped from
$\bm x$. Unlike the original LIQE, which samples sub-images only
at the native resolution, our model employs a multi-scale image
representation. Specifically, we construct a three-level image pyramid consisting
of the original image, a version downsampled by a factor of $0.75$,
and a version whose shorter side is resized to $224$ pixels while
preserving its aspect ratio. Sub-images are then cropped from all
three levels. On the textual side, the language encoder produces
the embedding matrix
$\bm G\in\mathbb R^{|\mathcal V|\times L}$
from the quality-related textual templates.

Following LIQE, we compute the cosine similarity between the
embedding of the $u$-th sub-image,
$\bm F_{u\bullet}(\bm x)$,
and the embedding of the $v$-th textual template,
$\bm G_{v\bullet}$.
Averaging these similarities over the $U$ sub-images gives
the image-level correspondence score:
\begin{align}\label{eq:logit}
\mathrm{logit}_{\bm w}(v|\bm x)
=
\frac{1}{U}
\sum_{u=1}^{U}
\frac{
\bm F_{u\bullet}(\bm x)
\bm G_{v\bullet}^{\intercal}
}{
\lVert\bm F_{u\bullet}(\bm x)\rVert_2
\lVert\bm G_{v\bullet}\rVert_2
}.
\end{align}
We then apply a softmax with learnable temperature $\tau$ to obtain
the probability distribution over the quality levels:
\begin{align}\label{eq:jointp}
p_{\bm w}(v|\bm x)
=
\frac{
\exp\!\left(
\mathrm{logit}_{\bm w}(v|\bm x)/\tau
\right)
}{
\displaystyle
\sum_{v'\in\mathcal V}
\exp\!\left(
\mathrm{logit}_{\bm w}(v'|\bm x)/\tau
\right)
}.
\end{align}
Substituting $p_{\bm w}(v|\bm x)$ into
Eq.~\eqref{eq:stl} yields the predicted quality score
$q_{\bm w}(\bm x)$. Following LIQE and UNIQUE, we train the model
using pairwise learning to rank across multiple IQA datasets. For an image pair $(\bm x,\bm y)$ from the same dataset, we define
the binary preference label:
\begin{align}\label{eq:bgt}
p(\bm x,\bm y)
=
\begin{cases}
1, & \text{if } q(\bm x)\geq q(\bm y),\\
0, & \text{otherwise}.
\end{cases}
\end{align}
Under the Thurstone model~\cite{thurstone1927law}, the predicted
probability that $\bm x$ has higher quality than $\bm y$ is
\begin{align}\label{eq:thurstone}
p_{\bm w}(\bm x,\bm y)
=
\Phi\left(
\frac{
q_{\bm w}(\bm x)-q_{\bm w}(\bm y)
}{
\sqrt{2}
}
\right).
\end{align}
Following~\cite{zhang2021uncertainty,zhang2023blind}, we optimize
the fidelity loss~\cite{tsai2007frank}:
\begin{align}\label{eq:fidelity}
\ell_{\bm w}(\bm x,\bm y)
={}&
1-
\sqrt{
p(\bm x,\bm y)\,
p_{\bm w}(\bm x,\bm y)
}
\nonumber\\
&-
\sqrt{
\bigl(1-p(\bm x,\bm y)\bigr)
\bigl(1-p_{\bm w}(\bm x,\bm y)\bigr)
}.
\end{align}
We call this model multi-scale LIQE (MS-LIQE).

\begin{table*}[t]
  \caption{Global ranking of eight NR-IQA models, where lower ranks indicate better performance. ``Multiple'' denotes training on the combined LIVE, CSIQ, BID, CLIVE, KonIQ-10k, and KADID-10k datasets. We define
$\Delta${Rank}$=${SRCC Rank}$-${MAP Rank};
thus, a positive value indicates a better rank under diffusion
latent MAP estimation.}
  \centering
    {\begin{tabular}{l|ccccc}
      \toprule
        NR-IQA Model & Training Set & \# Parameters (in Millions) & SRCC Rank (SRCC) & MAP Rank & $\Delta$ Rank \\
     \hline
        NIQE~\cite{mittal2013making} & -- & 0.001 & 8 (0.706)  & 8 & 0 \\
        CLIPIQA+~\cite{wang2023exploring} & KonIQ-10k  & 101.349 & 2 (0.855) & 6 & -4 \\
        PaQ-2-PiQ~\cite{ying2020from} & FLIVE & 11.704 & 5 (0.827) & 7 & -2 \\
        HyperIQA~\cite{su2020blindly} & KonIQ-10k & 27.375 & 7 (0.776) & 4 & 3  \\
        MUSIQ~\cite{ke2021musiq} & KonIQ-10k & 27.125 & 3 (0.853) & 2 & 1 \\
        DBCNN~\cite{zhang2020blind} & KonIQ-10k & 15.311 & 6 (0.789) & 5 & 1 \\
        UNIQUE~\cite{zhang2021uncertainty} & Multiple &22.322 & 4 (0.838) & 3 & 1 \\
        LIQE~\cite{zhang2023blind} & Multiple & 150.976 & 1 (0.881) & 1 & 0 \\
     \bottomrule
  \end{tabular}}
\label{tab:rank}
\end{table*}

\begin{figure*}[t]
    \centering
      \includegraphics[width=0.70\textwidth]{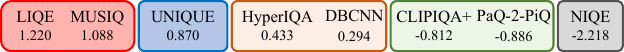}
    \caption{Global ranking scores of eight NR-IQA models under diffusion latent MAP estimation, where higher scores indicate better performance. Models enclosed by the same colored box are statistically indistinguishable according to a two-tailed $z$-test at the
$5\%$ significance level.  }\label{fig:results}
\end{figure*}

\section{Experiments}
\label{sec:exp}
In this section, we describe the experimental setups and compare
the eight NR-IQA models within the proposed framework. We then
evaluate MS-LIQE, conduct ablation studies, and demonstrate the
use of our framework as a post-enhancement step.

\subsection{Experimental Setups}
\label{subsec:selection}
\noindent\textbf{Image Selection.}  We construct the image set $\mathcal{X}$ by selecting $500$ photographic images from three widely used IQA datasets: CLIVE~\cite{ghadiyaram2016massive}, KonIQ-10k~\cite{hosu2020koniq}, and SPAQ~\cite{fang2020perceptual}. In contrast to simulated enhancement tasks with simple, clear goals (\eg, removing Gaussian noise and blur), our focus lies in enhancing images afflicted by mixed, heterogeneous distortions~\cite{ghadiyaram2016massive}. Such
 degradations, typically introduced during image acquisition and subsequent processing, rarely conform to simple categorical definitions
 (see Fig.~\ref{fig:ref_imgs}). To ensure input consistency, all images are resized to $512 \times 512$ pixels via bicubic interpolation followed by center cropping. We exclude images of extremely low quality or lacking a clear foreground subject, as they offer limited potential for meaningful quality discrimination or perceptual improvement.

\noindent\textbf{Model Selection.} We evaluate eight representative NR-IQA models, chosen to capture diverse design choices (see Table~\ref{tab:rank}): 1) {NIQE}~\cite{mittal2013making}, a knowledge-driven and training-free method that measures the distance between a test image and pristine natural images in a quality-aware feature space; 2) {DBCNN}~\cite{zhang2020blind}, a dual-stream architecture that combines features learned from
synthetic distortion classification with VGG-16
features~\cite{simonyan2014very} for authentic distortions; 3)  {HyperIQA}~\cite{su2020blindly}, which adopts a hyper-network for self-adaptive NR-IQA; 4) {PaQ-2-PiQ}~\cite{ying2020from}, a region-based model that produces both global quality scores and spatial quality maps; 5) {UNIQUE}~\cite{zhang2021uncertainty}, an uncertainty-aware method that unifies synthetic and authentic distortion assessment through pairwise ranking; 6) {MUSIQ}~\cite{ke2021musiq}, a multi-scale vision Transformer~\cite{dosovitskiy2021image} using hash-based space and scale embeddings; 7) {CLIPIQA+}~\cite{wang2023exploring}, which adapts pretrained CLIP~\cite{radford2021learning} via prompt learning~\cite{zhou2022learning}; and 8) {LIQE}~\cite{zhang2023blind}, which further finetunes both CLIP encoders across multiple IQA datasets~\cite{zhang2021uncertainty} while introducing scene classification and distortion type identification as auxiliary tasks. \textcolor{red}{
For each IQA dataset used to train the NR-IQA models, we adopt
training, validation, and test splits with a ratio of $7{:}1{:}2$.
The models are trained exclusively on the training subsets,
whereas the 500 candidate images are drawn only from the validation
and test subsets, thereby preventing content overlap with the
training data.
}

  \begin{figure*}[t]
    \centering
      \subfloat[\scriptsize Initial]{\includegraphics[width=0.24\textwidth]{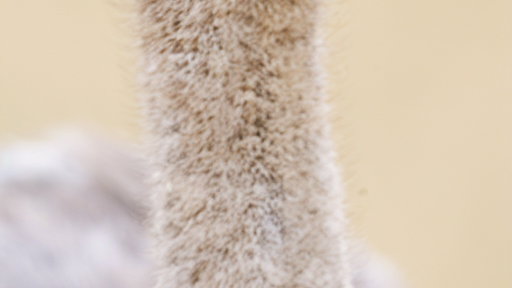}}\hskip.1em
      \vspace{-0.75mm}
    \subfloat[\scriptsize Initial]{\includegraphics[width=0.24\textwidth]{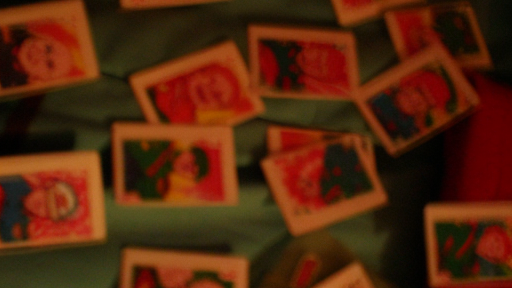}}\hskip.1em
    \vspace{-0.75mm}
    \subfloat[\scriptsize Initial]{\includegraphics[width=0.24\textwidth]{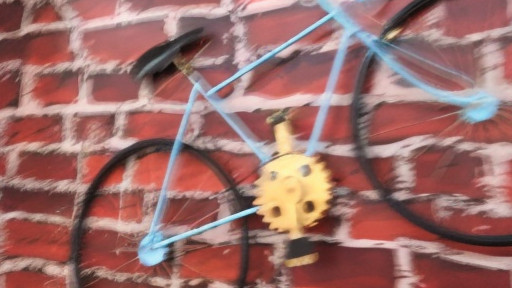}}\hskip.1em
    \vspace{-0.75mm}
      \subfloat[\scriptsize Initial]{\includegraphics[width=0.24\textwidth]{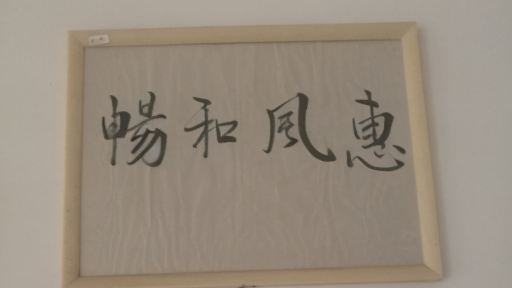}}\hskip.1em
      \vspace{-0.75mm}
      
    \subfloat[\scriptsize LIQE]{\includegraphics[width=0.24\textwidth]{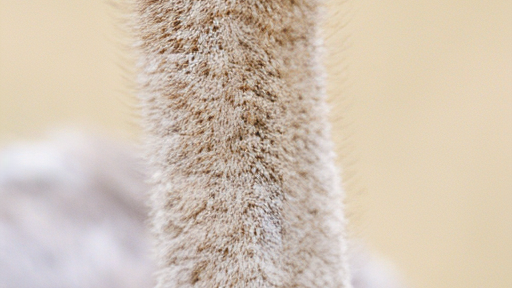}}\hskip.1em
    \vspace{-0.75mm}
      \subfloat[\scriptsize LIQE]{\includegraphics[width=0.24\textwidth]{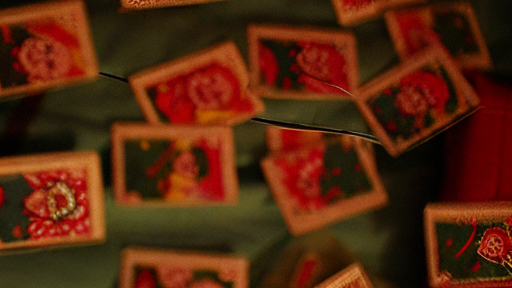}}\hskip.1em
      \vspace{-0.75mm}
    \subfloat[\scriptsize UNIQUE]{\includegraphics[width=0.24\textwidth]{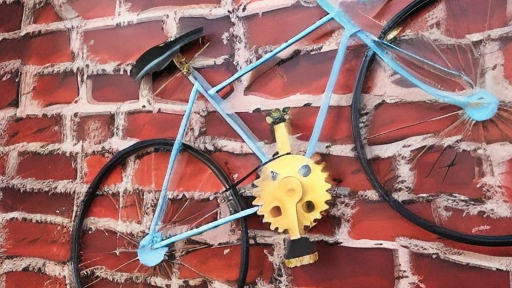}}\hskip.1em
    \vspace{-0.75mm}
    \subfloat[\scriptsize NIQE]{\includegraphics[width=0.24\textwidth]{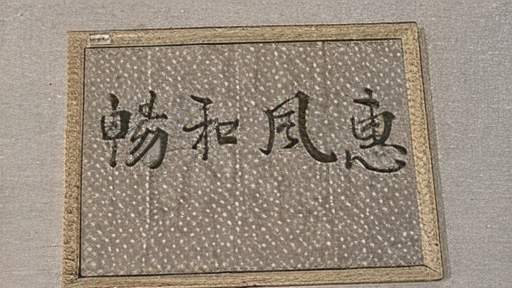}}\hskip.1em
    \vspace{-0.75mm}
    
      \subfloat[\scriptsize MUSIQ]{\includegraphics[width=0.24\textwidth]{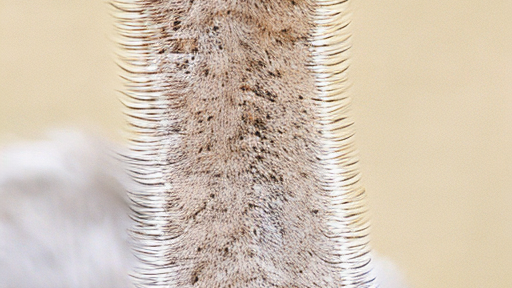}}\hskip.1em
    \subfloat[\scriptsize MUSIQ]{\includegraphics[width=0.24\textwidth]{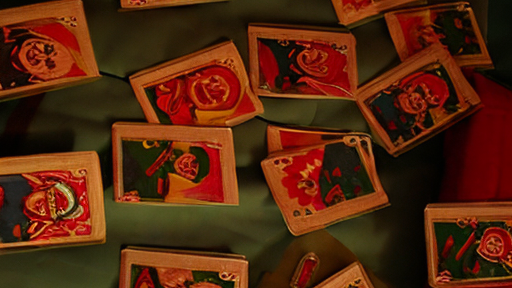}}\hskip.1em
    \subfloat[\scriptsize CLIPIQA+]{\includegraphics[width=0.24\textwidth]{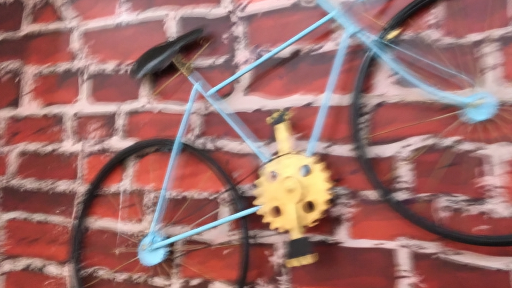}}\hskip.1em
      \subfloat[\scriptsize CLIPIQA+]{\includegraphics[width=0.24\textwidth]{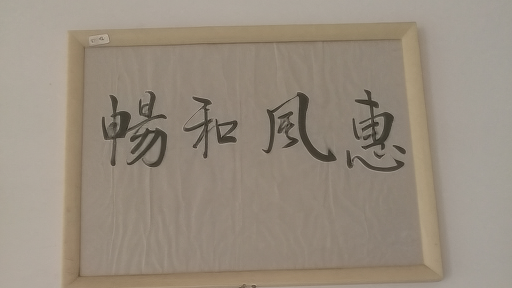}}\hskip.1em
    \caption{Visual comparison of MAP-refined images associated with different pairs of NR-IQA models.}\label{fig:qa_qualit}
\end{figure*}

\begin{figure*}[t]
    \centering
    \subfloat[\scriptsize Initial ]{\includegraphics[width=0.24\textwidth]{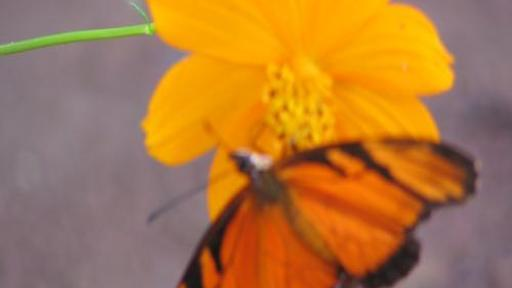}}\hskip.1em 
     \vspace{-0.75mm}
    \subfloat[\scriptsize Initial ]{\includegraphics[width=0.24\textwidth]{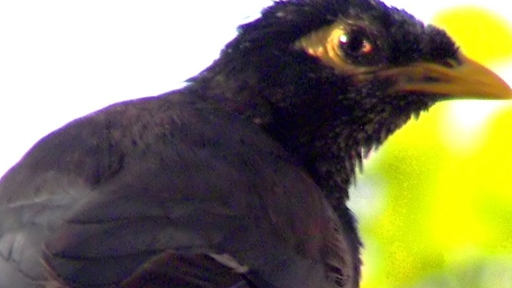}}\hskip.1em 
     \vspace{-0.75mm}
    \subfloat[\scriptsize Initial ]
    {\includegraphics[width=0.24\textwidth]{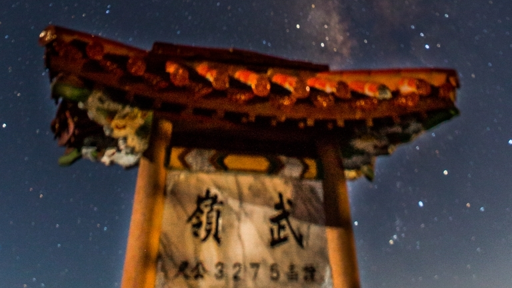}}\hskip.1em 
     \vspace{-0.75mm}
    \subfloat[\scriptsize Initial ]
    {\includegraphics[width=0.24\textwidth]{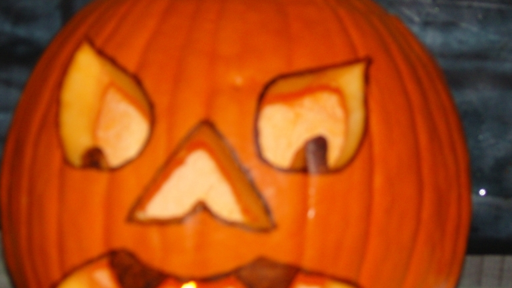}}\hskip.1em 
     \vspace{-0.75mm}
    \subfloat[\scriptsize Vanilla]{\includegraphics[width=0.24\textwidth]{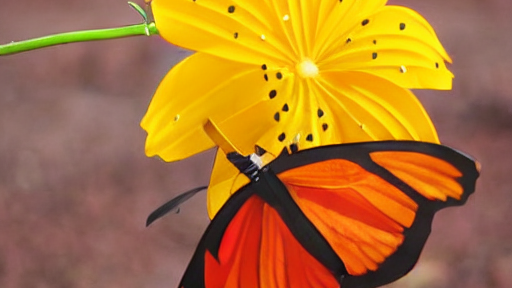}}\hskip.1em 
     \vspace{-0.75mm}
    \subfloat[\scriptsize Vanilla]{\includegraphics[width=0.24\textwidth]{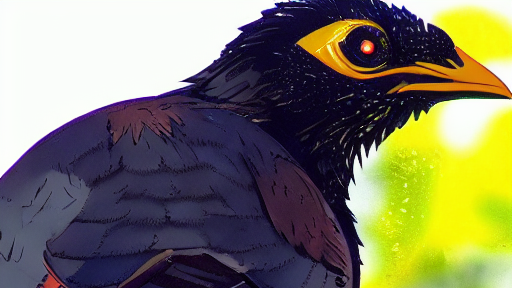}}\hskip.1em 
     \vspace{-0.75mm}
    \subfloat[\scriptsize Vanilla]
    {\includegraphics[width=0.24\textwidth]{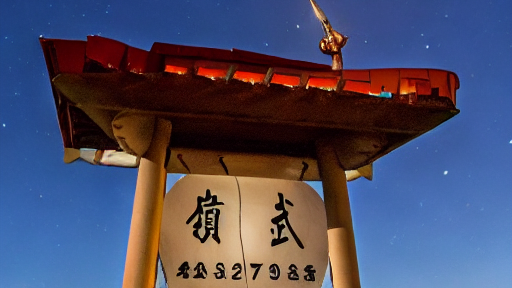}}\hskip.1em  
     \vspace{-0.75mm}
    \subfloat[\scriptsize Vanilla]
    {\includegraphics[width=0.24\textwidth]{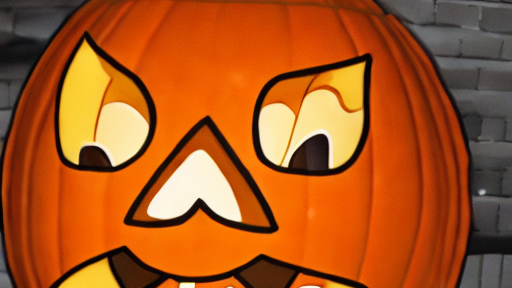}}\hskip.1em 
     \vspace{-0.75mm}
    \subfloat[\scriptsize ControlNet-augmented]{\includegraphics[width=0.24\textwidth]{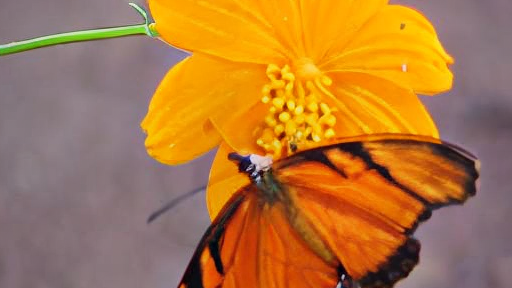}}\hskip.1em 
    \subfloat[\scriptsize ControlNet-augmented]{\includegraphics[width=0.24\textwidth]{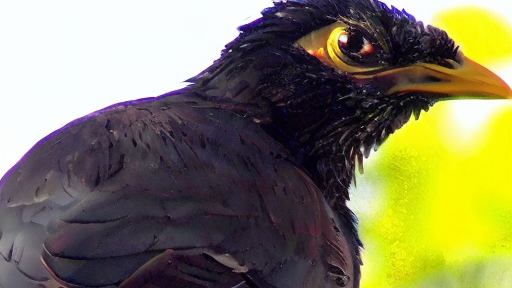}}\hskip.1em 
    \subfloat[\scriptsize ControlNet-augmented]
    {\includegraphics[width=0.24\textwidth]{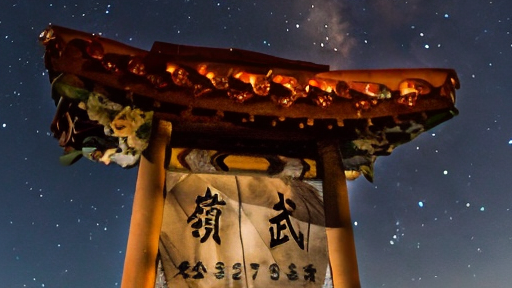}}\hskip.1em  
    \subfloat[\scriptsize ControlNet-augmented]
    {\includegraphics[width=0.24\textwidth]{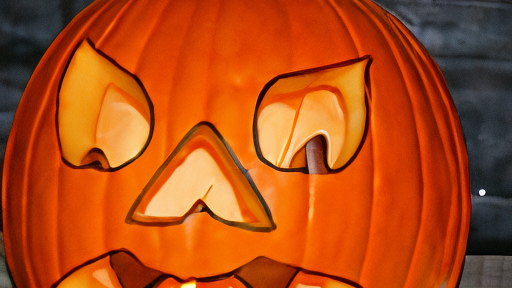}}\hskip.1em 
    \caption{Visual comparison of MAP-refined images using vanilla and ControlNet-augmented Stable Diffusion 2.0.}  
\label{fig:failure_cases}
\end{figure*}

 \begin{figure}[t]
    \centering
    \subfloat[]{\includegraphics[width=0.16\textwidth]{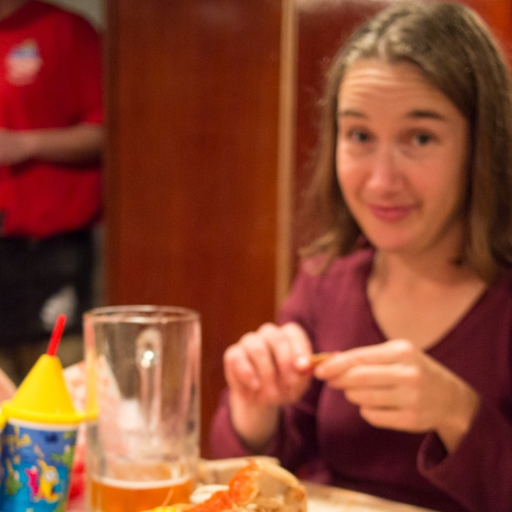}}\hskip.1em
      \subfloat[]{\includegraphics[width=0.16\textwidth]{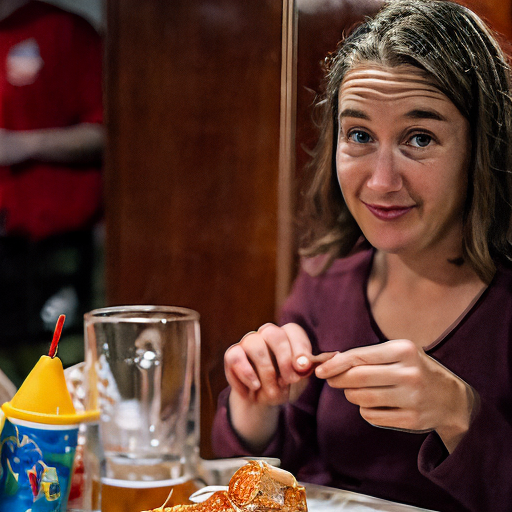}}\hskip.1em
    \subfloat[]{\includegraphics[width=0.16\textwidth]{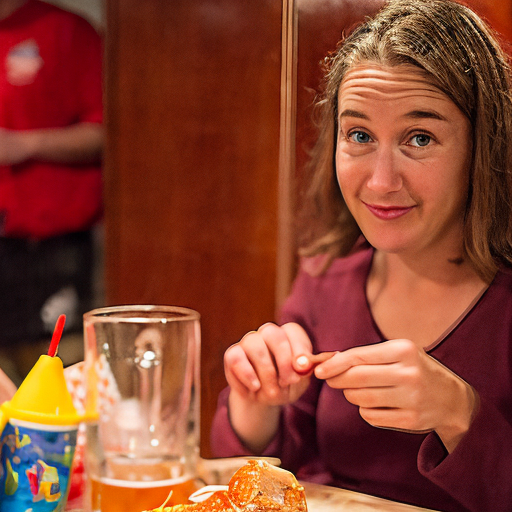}}
    \caption{\textbf{(a)} Initial image with mixed, realistic degradations. \textbf{(b)} \textcolor{red}{Enhancement output} by direct diffusion latent optimization of LIQE~\cite{zhang2023blind} (equivalent to diffusion latent MAP estimation with $\lambda = \infty$), which results in an overall color appearance that undesirably deviates from (a). \textbf{(c)} \textcolor{red}{Enhancement output} by the proposed diffusion latent MAP estimation using the same LIQE~\cite{zhang2023blind}. }\label{fig:example2}
\end{figure}

\noindent\textbf{Image Fidelity Measure.}
The fidelity term $D_0(\cdot,\cdot)$ preserves the global content
and appearance of the initial image while allowing plausible local
details to emerge during enhancement. We compute the Euclidean
distance between the optimized and initial images after spatial
downsampling:
\begin{align}
D_0(\bm x,\bm x^\mathrm{init})
=
\left\lVert
\bm x_{\downarrow s}
-
\bm x^\mathrm{init}_{\downarrow s}
\right\rVert_2,
\end{align}
where ${\downarrow s}$ denotes bicubic downsampling by a factor of
$s$. \textcolor{red}{We use $s=32$ in all experiments.}

\noindent\textbf{Optimization Details.} 
Following the recommendations of the Video Quality Experts Group~\cite{video2003final},
 we calibrate the predictions of each
NR-IQA model using a parametric logistic function:
\begin{align}\label{eq:nonlinear}
q_{\bm w,\bm\xi}(\bm x)
&=
\mathrm{logistic}\bigl(q_{\bm w}(\bm x)\bigr)
\nonumber\\
&=
\frac{\xi_1-\xi_2}{
1+
\exp\left(
-\frac{
q_{\bm w}(\bm x)-\xi_3
}{
|\xi_4|
}
\right)
}
+\xi_2.
\end{align}
We fix $\xi_1=1$ and $\xi_2=0$ to constrain the calibrated scores
to $[0,1]$. \textcolor{red}{For each NR-IQA model, $\xi_3$ and $\xi_4$ are fitted
independently by nonlinear least squares using its predictions and
the corresponding normalized MOSs on LIVE~\cite{sheikh2006statistical}.
They are initialized to the mean and standard deviation of the raw
predictions, respectively.}
We use $q_{\bm w,\bm\xi}$ in place of
$q_{\bm w}$ in Eq.~\eqref{eq:mapdl} and Algorithm~\ref{algoe}.
This calibration places the models on a common perceptual scale and
allows the same trade-off parameter $\lambda$ to be used across
all models.

For the diffusion backbone, we integrate \textcolor{red}{Stable Diffusion 2.0
Base~\cite{rombach2022high}} into EDICT and augment it with
ControlNet~\cite{zhang2023adding} to condition generation explicitly
on the input image~\cite{wu2024seesr}. \textcolor{red}{Because Stable Diffusion
operates in the VAE latent space, EDICT is bijective only between
the VAE latent and the corresponding diffusion latent; the lossy
pixel-latent mapping of the VAE lies outside this guarantee.} The ControlNet uses the standard full-size architecture matched to
the diffusion backbone and is pretrained on low- and high-quality
image pairs~\cite{karras2019style,li2023lsdir}. Both Stable Diffusion
and ControlNet remain frozen during optimization. We use a null text
prompt and set the ControlNet conditioning scale to $1.0$.

Following~\cite{wallace2023edict}, we use $T=20$ diffusion steps
and set the EDICT mixing coefficient to
$\eta=0.93^{50/T}$ in Eqs.~\eqref{eq:ii3} and~\eqref{eq:ii4}.
We set the MAP trade-off parameter to $\lambda=0.01$ and perform
at most $\mathrm{MaxIter}=40$ optimization iterations using a
learning rate of $\alpha_{\mathrm{lr}}=2$ and momentum
$\rho=0.9$. For image selection in Eq.~\eqref{eq:mad}, we set
$\gamma=0.1$. \textcolor{red}{The same configuration is used for every NR-IQA
model without model-specific tuning.}

\noindent\textbf{Psychophysical Testing Details.}
The experiment was conducted in a controlled office environment
with an ambient illuminance of approximately $200\,\mathrm{lux}$
and nonreflective ceilings, walls, and floors. The display was
calibrated to a peak luminance of
$200\,\mathrm{cd}/\mathrm{m}^2$.
All image pairs were presented at their native resolution, with a
viewing distance of $32$ pixels per degree of visual angle.

We used a 2AFC protocol in which
participants selected the output with higher perceived quality.
The study involved $25$ participants
($13$ male and $12$ female, aged $23$-$36$ years)
with normal or corrected-to-normal vision. Before testing,
participants completed a training session and could take breaks
at any time to mitigate fatigue.

\textcolor{red}{For each of the $\binom{8}{2}=28$ NR-IQA model pairs, we selected
$20$ image pairs, yielding $560$ comparisons per participant. We added
$20$ repeated trials to assess intra-participant consistency and
retained participants whose responses agreed on at least $80\%$
of these trials. All $25$ participants satisfied this criterion.}

\textcolor{red}{For each participant, the presentation order and left-right
arrangement of the outputs were independently randomized.
Comparisons from different model pairs were interleaved, and model
identities were concealed. Each selected comparison was evaluated
once by each participant. The basic unit of statistical analysis
was one binary response from one participant to one selected
comparison.}

\textbf{MS-LIQE Training Details.} MS-LIQE is trained using AdamW~\cite{LoshchilovH19} with a decoupled weight decay of $10^{-3}$ and a mini-batch size of $48$. The initial learning rate is set to $5\times10^{-6}$ and scheduled following a cosine annealing rule~\cite{LoshchilovH17}. We optimize MS-LIQE for $80$ epochs over a combined dataset, comprising LIVE~\cite{sheikh2006statistical}, CSIQ~\cite{larson2010most}, BID~\cite{ciancio2011no}, CLIVE~\cite{ghadiyaram2016massive}, KonIQ-10k~\cite{hosu2020koniq}, and KADID-10k~\cite{lin2019kadid}.

  \begin{figure*}[!t]
    \centering
      \subfloat[\scriptsize Initial]{\includegraphics[width=0.24\textwidth]{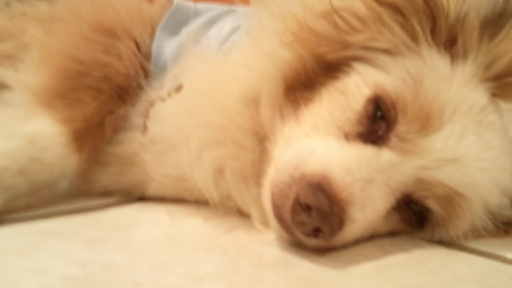}}\hskip.1em
      \vspace{-0.75mm}
    \subfloat[\scriptsize Initial]{\includegraphics[width=0.24\textwidth]{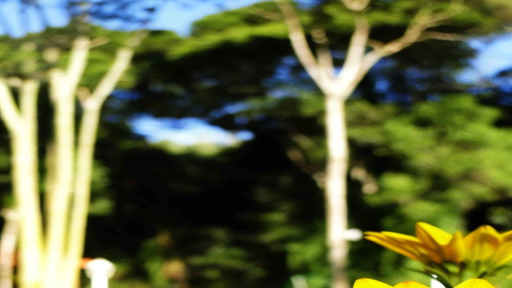}}\hskip.1em
    \vspace{-0.75mm}
    \subfloat[\scriptsize Initial]{\includegraphics[width=0.24\textwidth]{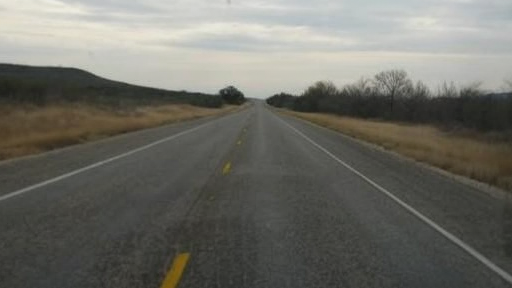}}\hskip.1em
    \vspace{-0.75mm}
      \subfloat[\scriptsize Initial]{\includegraphics[width=0.24\textwidth]{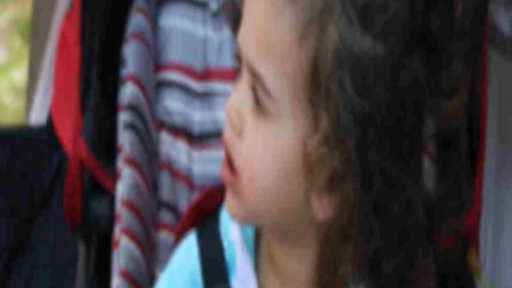}}\hskip.1em
      \vspace{-0.75mm}
      
    \subfloat[\scriptsize LIQE]{\includegraphics[width=0.24\textwidth]{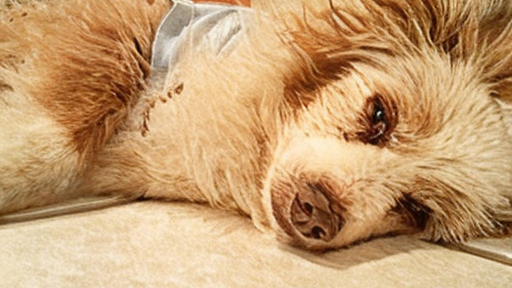}}\hskip.1em
    \vspace{-0.75mm}
    \subfloat[\scriptsize LIQE]{\includegraphics[width=0.24\textwidth]{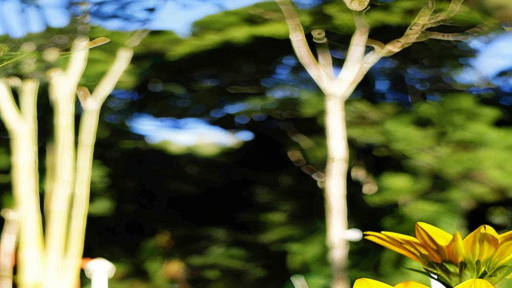}}\hskip.1em
      \vspace{-0.75mm}
    \subfloat[\scriptsize MUSIQ]{\includegraphics[width=0.24\textwidth]{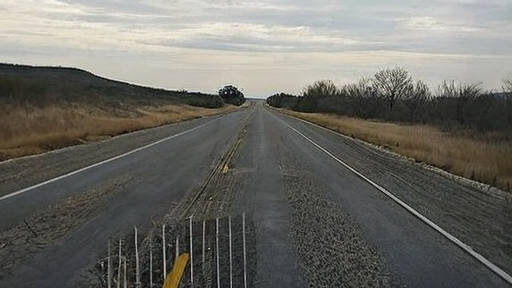}}\hskip.1em
    \vspace{-0.75mm}
    \subfloat[\scriptsize UNIQUE]{\includegraphics[width=0.24\textwidth]{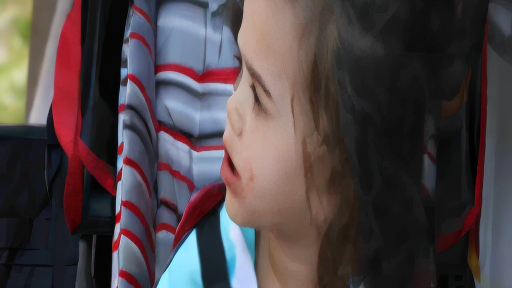}}\hskip.1em
    \vspace{-0.75mm}
    
      \subfloat[\scriptsize MS-LIQE]{\includegraphics[width=0.24\textwidth]{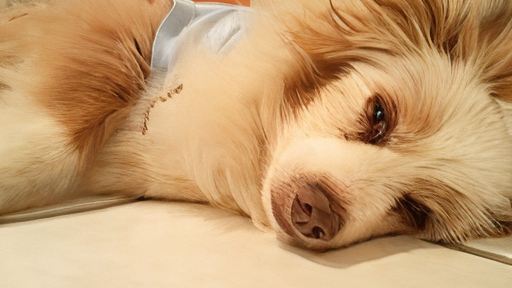}}\hskip.1em
    \subfloat[\scriptsize MS-LIQE]{\includegraphics[width=0.24\textwidth]{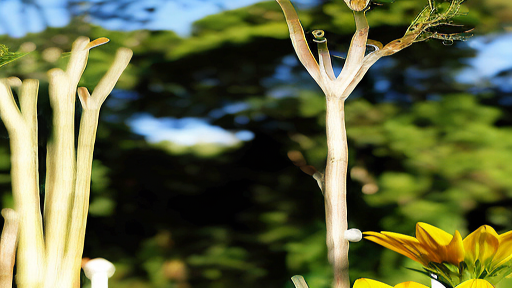}}\hskip.1em
    \subfloat[\scriptsize MS-LIQE]{\includegraphics[width=0.24\textwidth]{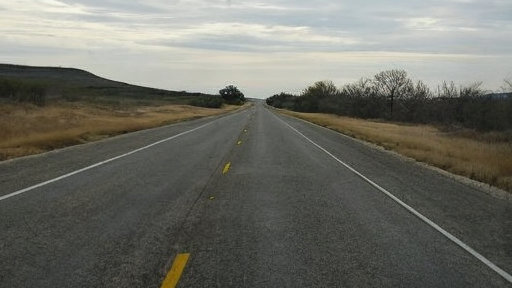}}\hskip.1em
      \subfloat[\scriptsize MS-LIQE]{\includegraphics[width=0.24\textwidth]{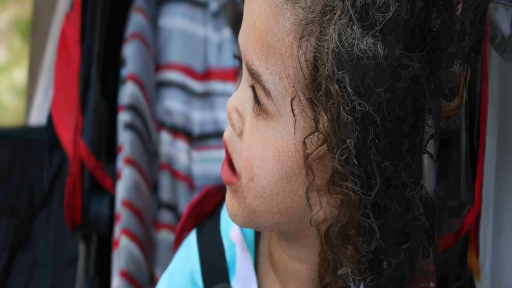}}\hskip.1em
    \caption{Visual comparison of MAP-refined images guided by MS-LIQE and
its three predecessor models: LIQE, MUSIQ, and UNIQUE.}\label{fig:qa_qualit2}
\end{figure*}

\subsection{Model Comparison Results}
\label{subsec:results}
\noindent\textbf{Quantitative Results.}
We estimate the global ranking scores of the eight NR-IQA models
using Eq.~\eqref{eq:jod}. \textcolor{red}{Their uncertainty and covariance are
estimated from $500$ subject-level bootstrap resamples. For each
model pair, we perform a two-sided $z$-test~\cite{david1963method}
of
$H_0:\mu_i-\mu_j=0$.}
Models for which the null hypothesis cannot be rejected at the
$5\%$ significance level are considered statistically
indistinguishable.

Fig.~\ref{fig:results} presents the estimated ranking scores,
with each colored box grouping models that are statistically
indistinguishable at the $5\%$ significance level. These results
support four main observations. {First}, LIQE~\cite{zhang2023blind} and UNIQUE~\cite{zhang2021uncertainty} significantly outperform most competing methods, with the exception of MUSIQ~\cite{ke2021musiq}, underscoring the effectiveness of ranking-based quality assessment across multiple datasets. The performance disparity between LIQE and UNIQUE can largely be explained by two factors: 1) the choice of backbone networks and 2) the scale of pretraining data. LIQE is built upon the CLIP model with more than $150$ million parameters, trained on $400$ million image-text pairs, whereas UNIQUE employs a ResNet-34 variant with approximately $22$ million parameters, trained on $1.28$ million images from ImageNet~\cite{deng2009imagenet}. {Second}, although MUSIQ, DBCNN~\cite{zhang2020blind}, and HyperIQA~\cite{su2020blindly} share comparable model capacity and training dataset (see Table~\ref{tab:rank}), MUSIQ significantly outperforms the latter two. This advantage may arise from its multi-scale image representations, which better align with the characteristics of human visual perception. {Third}, CLIPIQA+~\cite{wang2023exploring} exhibits relatively weak performance despite sharing the CLIP backbone with the top-performing LIQE. A closer examination reveals that CLIPIQA+ relies on prompt learning~\cite{zhou2022learning}, while keeping model weights fixed, which likely limits its ability to learn or capture quality-aware representations. {Fourth}, the knowledge-driven NIQE~\cite{mittal2013making}, though designed for arbitrary distortions, performs poorly under our diffusion latent MAP estimation framework. This highlights the difficulty of manually designing features to capture the intricate interactions between natural image statistics and visual quality.

 We further compare the ranking results produced by the proposed diffusion latent MAP estimation framework with those obtained using Spearman's rank correlation coefficient (SRCC) on the SPAQ dataset~\cite{fang2020perceptual}. Importantly, all NR-IQA models are evaluated in a cross-dataset setting (\ie, not trained on SPAQ). A key observation from Table~\ref{tab:rank} is that superior SRCC performance does not necessarily imply stronger performance under diffusion latent MAP estimation. \textcolor{red}{One plausible explanation for this discrepancy is shortcut
learning~\cite{geirhos2020shortcut}. NR-IQA models with comparable
prediction accuracy may rely on different dataset-specific cues
rather than genuinely quality-aware representations. Such shortcuts
may remain hidden in correlation-based evaluations but become
apparent when the models are used as optimization objectives.} Overall, these findings highlight the effectiveness of our framework in discriminating the relative performance of NR-IQA models through analysis-by-synthesis, thereby offering a complementary perspective to correlation-based evaluation.

\noindent\textbf{Qualitative Results.}
We show representative MAP-refined images generated from different pairs of NR-IQA models in Fig.~\ref{fig:qa_qualit}, automatically identified using Eq.~\eqref{eq:mad}.
We begin by examining LIQE and MUSIQ, the two leading NR-IQA models under diffusion latent MAP estimation. As shown in Fig.~\ref{fig:qa_qualit}(e), LIQE successfully removes blur near the animal's tail, restoring natural fur textures. In contrast, MUSIQ introduces hallucination artifacts, producing overly magnified and unnatural fine hairs along the tail boundary (see Fig.~\ref{fig:qa_qualit}(i)). Interestingly, MUSIQ performs better than
LIQE when enhancing stamps (see Figs.~\ref{fig:qa_qualit}(f) and (j)), suggesting the complementary strengths of the two models. 

We next turn to CLIPIQA+, another CLIP-based NR-IQA model, which exhibits a relatively conservative strategy, resulting in minimal perceptually meaningful pixel changes (see Figs.~\ref{fig:qa_qualit}(k) and (l)). Such behavior can be advantageous or detrimental depending on the input quality. For high-quality inputs, such as Fig.~\ref{fig:qa_qualit}(d), avoiding pixel modifications is preferable to introducing distortions, as exemplified by NIQE in Fig.~\ref{fig:qa_qualit}(h). Conversely, for low-quality inputs, more aggressive enhancement strategies, such as those applied by UNIQUE in Fig.~\ref{fig:qa_qualit}(g), generally yield more perceptually favorable results.

\subsection{Ablation Studies}\label{subsec:ablation}
To assess the effectiveness of the chosen diffusion backbone, we benchmark it against vanilla Stable Diffusion 2.0 without ControlNet~\cite{zhang2023adding}. A close visual inspection of the MAP-refined images in Fig.~\ref{fig:failure_cases} reveals a key limitation of the vanilla model: its tendency to generate hallucinated, over-smoothed, and cartoon-like content. This weakness is rooted in the inherent bias and stochasticity of diffusion models in image generation~\cite{kadkhodaie2021stochastic}. In contrast, 
the ControlNet-augmented variant leverages the input image as conditioning, thus imposing an implicit fidelity constraint that complements the explicit fidelity term in Eq.~\eqref{eq:mapdl}. Together, these dual mechanisms achieve a more favorable balance between fidelity and quality.

We next assess the contribution of the fidelity term by comparing
diffusion latent MAP estimation with direct diffusion-latent
optimization~\cite{wallace2023end}. As shown in
Figs.~\ref{fig:example2}(b) and (c), both approaches improve the
perceptual quality of the input. However, the MAP formulation
better preserves the color appearance of the initial image,
demonstrating the benefit of the explicit fidelity term.

\subsection{Improved Model Results}\label{subsec:finetune2}
We compare MS-LIQE against its predecessors---LIQE, MUSIQ, and UNIQUE---within the proposed diffusion latent MAP estimation framework. To compare their enhancement performance quantitatively, we apply
the debiased psychophysical procedure described in
Sec.~\ref{subsec:subjective}. We select $20$ informative image pairs
for each of the three comparisons between MS-LIQE and LIQE, MUSIQ,
or UNIQUE, yielding $60$ pairs in total. The outputs guided by
MS-LIQE receive $76.4\%$, $78.6\%$, and $82.5\%$ of the preference
votes against LIQE, MUSIQ, and UNIQUE, respectively.

MS-LIQE consistently produces more natural and realistic results than the compared methods. As shown in Fig.~\ref{fig:qa_qualit2}(i), it suppresses noisy, grainy artifacts in fur and background regions more effectively than LIQE, yielding cleaner textures and tones. In Fig.~\ref{fig:qa_qualit2}(j), it also better preserves depth of field by clearly separating foreground, midground, and background, whereas LIQE merges the latter two. This may stem from the multi-scale representations in MS-LIQE, which improve sensitivity to structures at different distances. Compared with MUSIQ, MS-LIQE generates clearer outputs with fewer unnatural artifacts, such as those visible in the lower road region of Fig.~\ref{fig:qa_qualit2}(g). Compared with UNIQUE, it produces more refined perceptual results with better sharpness and clarity, especially in hair textures, while avoiding the overaggressive deblurring that can make UNIQUE's outputs look less natural.

\begin{figure*}[t]
    \centering
    \subfloat[\scriptsize Initial (SR) ]{\includegraphics[width=0.24\textwidth]{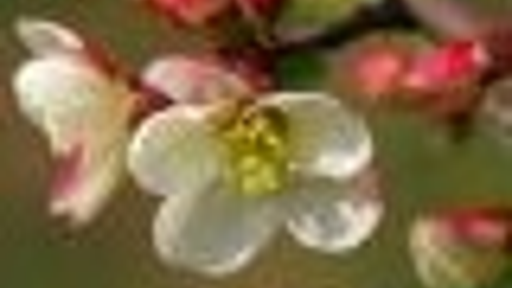}}\hskip.1em 
     \vspace{-0.75mm}
    \subfloat[\scriptsize Initial (SR) ]{\includegraphics[width=0.24\textwidth]{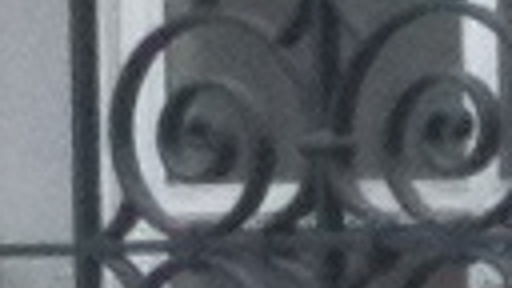}}\hskip.1em 
     \vspace{-0.75mm}
    \subfloat[\scriptsize Initial (LLIE) ]
    {\includegraphics[width=0.24\textwidth]{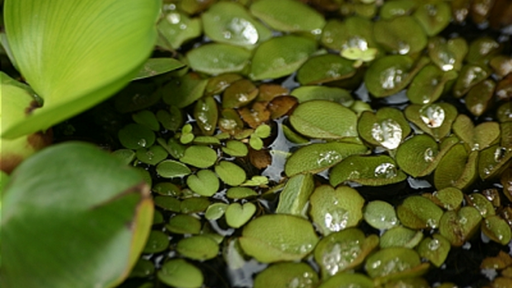}}\hskip.1em 
     \vspace{-0.75mm}
    \subfloat[\scriptsize Initial (FR)]
    {\includegraphics[width=0.24\textwidth]{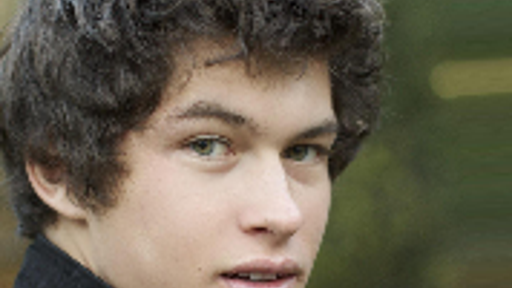}}\hskip.1em 
     \vspace{-0.75mm}
    \subfloat[\scriptsize SeeSR]{\includegraphics[width=0.24\textwidth]{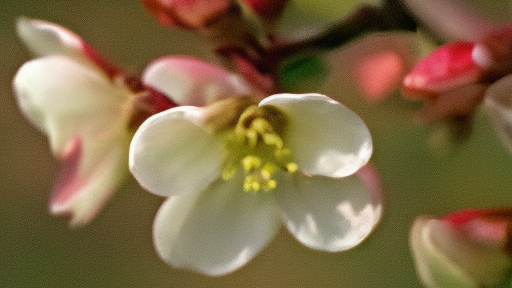}}\hskip.1em 
     \vspace{-0.75mm}
    \subfloat[\scriptsize PiSA-SR]{\includegraphics[width=0.24\textwidth]{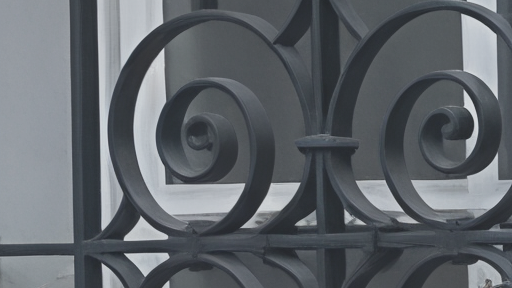}}\hskip.1em 
     \vspace{-0.75mm}
    \subfloat[\scriptsize HVI-CIDNet]
    {\includegraphics[width=0.24\textwidth]{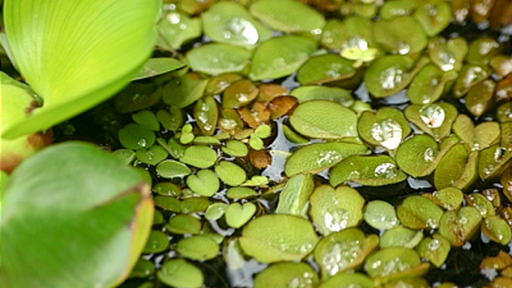}}\hskip.1em  
     \vspace{-0.75mm}
    \subfloat[\scriptsize SUPIR]
    {\includegraphics[width=0.24\textwidth]{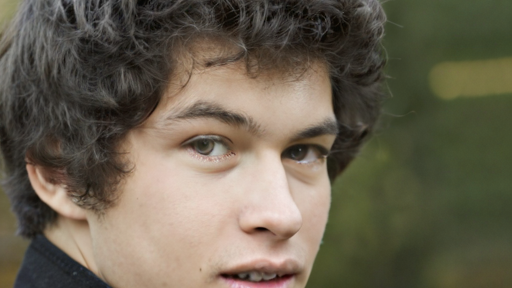}}\hskip.1em 
     \vspace{-0.75mm}
    \subfloat[\scriptsize SeeSR + Ours]{\includegraphics[width=0.24\textwidth]{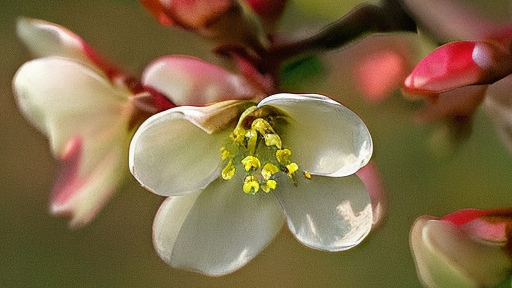}}\hskip.1em 
    \subfloat[\scriptsize PiSA-SR + Ours]{\includegraphics[width=0.24\textwidth]{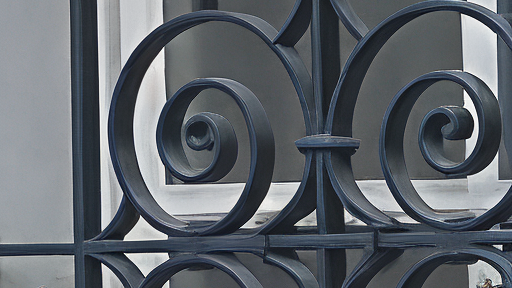}}\hskip.1em 
    \subfloat[\scriptsize HVI-CIDNet + Ours]
    {\includegraphics[width=0.24\textwidth]{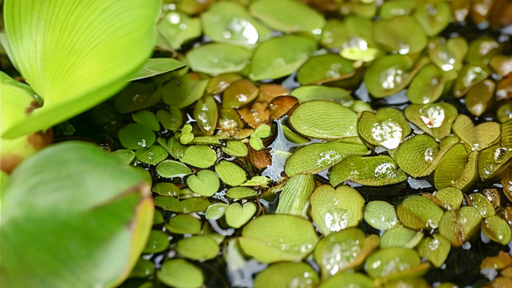}}\hskip.1em  
    \subfloat[\scriptsize SUPIR + Ours]
    {\includegraphics[width=0.24\textwidth]{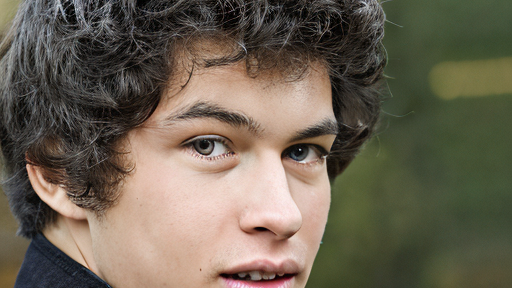}}\hskip.1em 
    \caption{Visual examples of the proposed post-enhancement method applied
    to task-specific enhancement outputs. From left to right, the
    columns show two super-resolution (SR) examples, one low-light
    image enhancement (LLIE) example, and one face restoration (FR)
    example.}  
\label{fig:postenhancement}
\end{figure*}

\begin{table*}[t]
\caption{
Quantitative comparison of the baseline and post-enhanced outputs
on super-resolution and face restoration.
Each entry reports {baseline / post-enhancement}, and higher values indicate
better performance for all metrics
}
\label{tab:sr_face_results}
\centering
\small
\setlength{\tabcolsep}{4.0pt}
\renewcommand{\arraystretch}{1.15}

\begin{tabular}{lccccccccc}
\toprule
\multirow{2}{*}{Method} 
& \multicolumn{3}{c}{Super Resolution on Real60} 
& \multicolumn{3}{c}{Super Resolution on DPED} 
& \multicolumn{3}{c}{Face Restoration} \\
\cmidrule(lr){2-4}\cmidrule(lr){5-7}\cmidrule(lr){8-10}
& MANIQA & Q-Align & VQ-R1
& MANIQA & Q-Align & VQ-R1
& MANIQA & Q-Align & VQ-R1 \\
\midrule

SeeSR
& \pair{0.655}{0.681} & \pair{4.46}{4.69} & \pair{4.65}{4.74}
& \pair{0.586}{0.635} & \pair{3.00}{3.62} & \pair{3.29}{3.79}
& \pair{0.633}{0.697} & \pair{3.96}{4.50} & \pair{4.42}{4.68} \\

SUPIR
& \pair{0.655}{0.671} & \pair{4.46}{4.77} & \pair{4.69}{4.77}
& \pair{0.574}{0.621} & \pair{2.70}{3.40} & \pair{3.06}{3.44}
& \pair{0.610}{0.686} & \pair{3.68}{4.46} & \pair{4.26}{4.63} \\

StableSR
& \pair{0.641}{0.668} & \pair{3.89}{4.53} & \pair{4.51}{4.73}
& \pair{0.536}{0.604} & \pair{2.43}{3.22} & \pair{2.72}{3.56}
& \pair{0.608}{0.671} & \pair{3.60}{4.38} & \pair{4.11}{4.60} \\

PiSA-SR
& \pair{0.688}{0.703} & \pair{4.40}{4.71} & \pair{4.70}{4.81}
& \pair{0.617}{0.658} & \pair{2.85}{3.44} & \pair{3.22}{3.73}
& \pair{0.663}{0.701} & \pair{4.11}{4.65} & \pair{4.51}{4.76} \\

\bottomrule
\end{tabular}
\end{table*}

\begin{table}[t]
\caption{Quantitative comparison of the baseline and post-enhanced outputs
on low-light image enhancement}
\label{tab:llie_results}
\centering
\small
\setlength{\tabcolsep}{3.2pt}
\renewcommand{\arraystretch}{1.05}

\begin{tabular}{lccc}
\toprule
Method & MANIQA & Q-Align & VQ-R1 \\
\midrule

QuadPrior
& \pair{0.634}{0.683} & \pair{3.71}{4.25} & \pair{4.12}{4.44} \\

HVI-CIDNet
& \pair{0.642}{0.664} & \pair{3.64}{4.18} & \pair{4.01}{4.37} \\

RetinexFormer
& \pair{0.570}{0.628} & \pair{3.25}{3.95} & \pair{3.63}{4.03} \\

ISP-UNet
& \pair{0.643}{0.678} & \pair{3.81}{4.29} & \pair{4.18}{4.43} \\

\bottomrule
\end{tabular}
\end{table}

\begin{table}[t]
\centering
\caption{\textcolor{red}{Percentage of trials in which the post-enhanced output is preferred
to the strongest competing baseline for three tasks}}
\label{tab:post_subjective}
\begin{tabular}{l l c}
\toprule
\textcolor{red}{Task}
& \textcolor{red}{Baseline}
& \textcolor{red}{Preference (\%)} \\
\midrule
\textcolor{red}{Super-Resolution on Real60}
& \textcolor{red}{PiSA-SR}
& \textcolor{red}{85.6} \\
\textcolor{red}{Super-Resolution on DPED}
& \textcolor{red}{SeeSR}
& \textcolor{red}{79.6} \\
\textcolor{red}{Face Restoration}
& \textcolor{red}{PiSA-SR}
& \textcolor{red}{88.0} \\
\textcolor{red}{Low-light Enhancement}
& \textcolor{red}{ISP-UNet}
& \textcolor{red}{76.8} \\
\bottomrule
\end{tabular}
\end{table}

\subsection{Post-enhancement Results}\label{subsec:refiner} The proposed diffusion latent MAP estimation framework is complementary to existing real-world image enhancers and can be used as a post-processing step to refine their outputs. \textcolor{red}{For all post-enhancement tasks, we work with MS-LIQE while keeping the diffusion backbone and optimization settings unchanged.} We validate this capability on three real-world image enhancement tasks: super-resolution on Real60~\cite{yu2024scaling} and DPED~\cite{ignatov2017dslr}, face restoration on the NTIRE Challenge~\cite{ntire2025_face_restoration}, and low-light image enhancement on the unpaired set~\cite{wang2026robust}. For super-resolution and face restoration, we experiment with four enhancers: SeeSR~\cite{wu2024seesr}, SUPIR~\cite{yu2024scaling}, StableSR~\cite{wang2024exploiting}, and PiSA-SR~\cite{sun2025pixel}. For low-light image enhancement, we use QuadPrior~\cite{Wang_2024_CVPR}, HVI-CIDNet~\cite{Yan_2025_CVPR}, RetinexFormer~\cite{Cai_2023_ICCV}, and ISP-UNet~\cite{wang2025towards} to obtain the initial enhancement results. We employ three NR-IQA models to quantify enhancement performance across all three tasks: MANIQA~\cite{yang2022maniqa}, Q-Align~\cite{wu2024q}, and VisualQuality-R1 (VQ-R1)~\cite{wu2025visualquality}. Their architectures and training data differ substantially from MS-LIQE, reducing the risk of evaluation bias and supporting their use as independent evaluation metrics. As shown in Tables~\ref{tab:sr_face_results} and~\ref{tab:llie_results}, our post-enhancement brings consistent gains over the corresponding baselines across all metrics on all three tasks. Furthermore, visual results in Fig.~\ref{fig:postenhancement} show that our method effectively restores fine details and suppresses artifacts present in the baseline outputs. 

\textcolor{red}{
To verify the perceptual improvements, we conduct an additional
psychophysical experiment. Based on
Tables~\ref{tab:sr_face_results} and~\ref{tab:llie_results},
we select the strongest competing baseline for each setting:
PiSA-SR on Real60, SeeSR on DPED, PiSA-SR for face restoration,
and ISP-UNet for low-light image enhancement. Following the debiased selection procedure, we identify $10$ informative comparison
pairs for each setting and evaluate them using the 2AFC protocol.
As shown in Table~\ref{tab:post_subjective}, participants consistently
prefer the post-enhanced outputs to the corresponding baseline
outputs. These results provide direct human-preference evidence
that the proposed framework is effective as a post-enhancement step.
}

\section{Conclusion and Discussion}\label{sec:conclusion}
We have introduced diffusion latent MAP estimation, which couples
existing NR-IQA models with an EDICT-based invertible diffusion
transform to optimize perceptual quality in the diffusion latent
space while preserving image fidelity. With the diffusion backbone,
fidelity measure, and optimization procedure held fixed, different
NR-IQA objectives produce distinct enhancement outputs, enabling
their optimization capabilities to be compared through
analysis-by-synthesis.

Our evaluation of eight representative NR-IQA models complements
conventional correlation-based assessment by examining how these
models behave as perceptual optimization objectives. Insights from
the strongest-performing models further guided the development of
MS-LIQE, which demonstrated improved enhancement performance within
the proposed framework. We also showed that the framework can serve
as a post-enhancement step for super-resolution, low-light image
enhancement, and face restoration.

Several limitations remain. First, the optimized outputs depend
jointly on the NR-IQA model, diffusion backbone, fidelity measure,
and optimization procedure. Although all components other than the
NR-IQA model are fixed in our experiments, the resulting rankings
should be interpreted as relative optimization capabilities within
the proposed framework rather than as framework-independent
assessments of model quality. Second, the framework depends on NR-IQA objectives that may be
susceptible to shortcut learning~\cite{geirhos2020shortcut}, whereby
higher predicted scores do not necessarily correspond to meaningful
perceptual improvements~\cite{weng2024rewardhack}. One promising
direction is to develop generative NR-IQA models that estimate the
quality-conditioned image distribution
$p(\bm x| q)$
rather than only predicting quality through
$p(q|\bm x)$.
\textcolor{red}{This formulation could align perceptual optimization more directly
with quality assessment while reducing reliance on auxiliary
generative components. Another important direction is to reduce
the computational cost of gradient evaluation through the full
diffusion process~\cite{zhang2024exact,wang2024belm,SongD0S23}.}

\ifCLASSOPTIONcompsoc
  \section*{Acknowledgments}
\else
  \section*{Acknowledgment}
\fi
This work was supported in part by the National Natural Science Foundation of China (62371283), the Guangdong and Hong Kong Universities `1+1+1' Joint Research Collaboration Scheme, and the Hong Kong ITC
Innovation and Technology Fund (9440379 and 9440390).

\bibliographystyle{IEEEtran}
\bibliography{Weixia}

\begin{IEEEbiography}[{\includegraphics[width=1in,height=1.25in,clip,keepaspectratio]{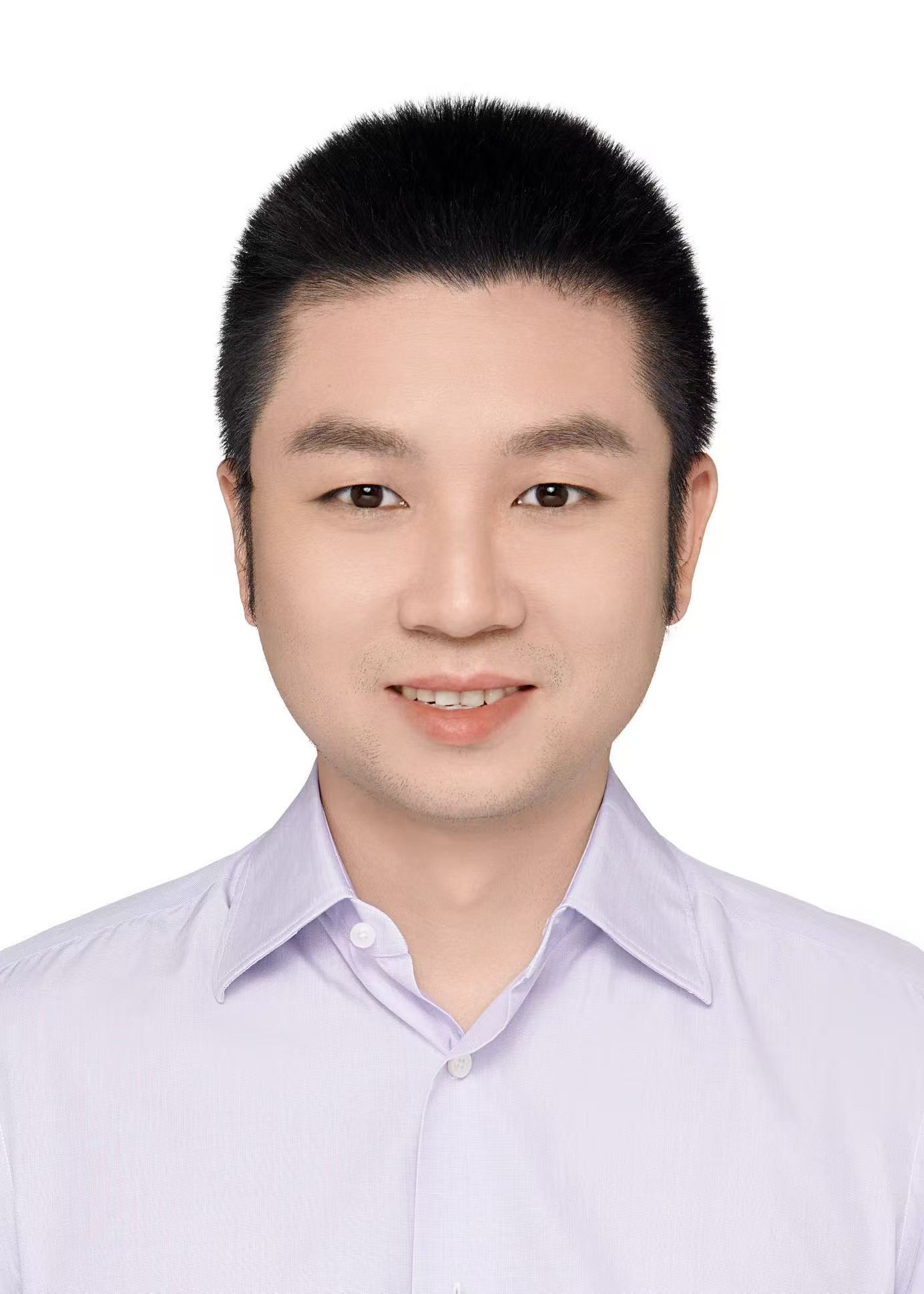}}]{Weixia Zhang}(Member, IEEE) is currently an Associate Research Scientist with the Institute of AI, School of Computer Science, Shanghai Jiao Tong University. He received the B.E. degree from Wuhan University, Wuhan, China, in 2011 and the M.S. degree in electrical and computer engineering from the University of Rochester, NY, USA, in 2013. He then received the Ph.D. degree from Wuhan University, Wuhan, China, in 2018. From 2018 to 2023, he was a Research Associate with Shanghai Jiao Tong University. His research interests include perceptual quality assessment and optimization. He currently serves on the Editorial Boards of IEEE Transactions on Image Processing.
\end{IEEEbiography}

\begin{IEEEbiography}[{\includegraphics[width=1in,height=1.25in,clip,keepaspectratio]{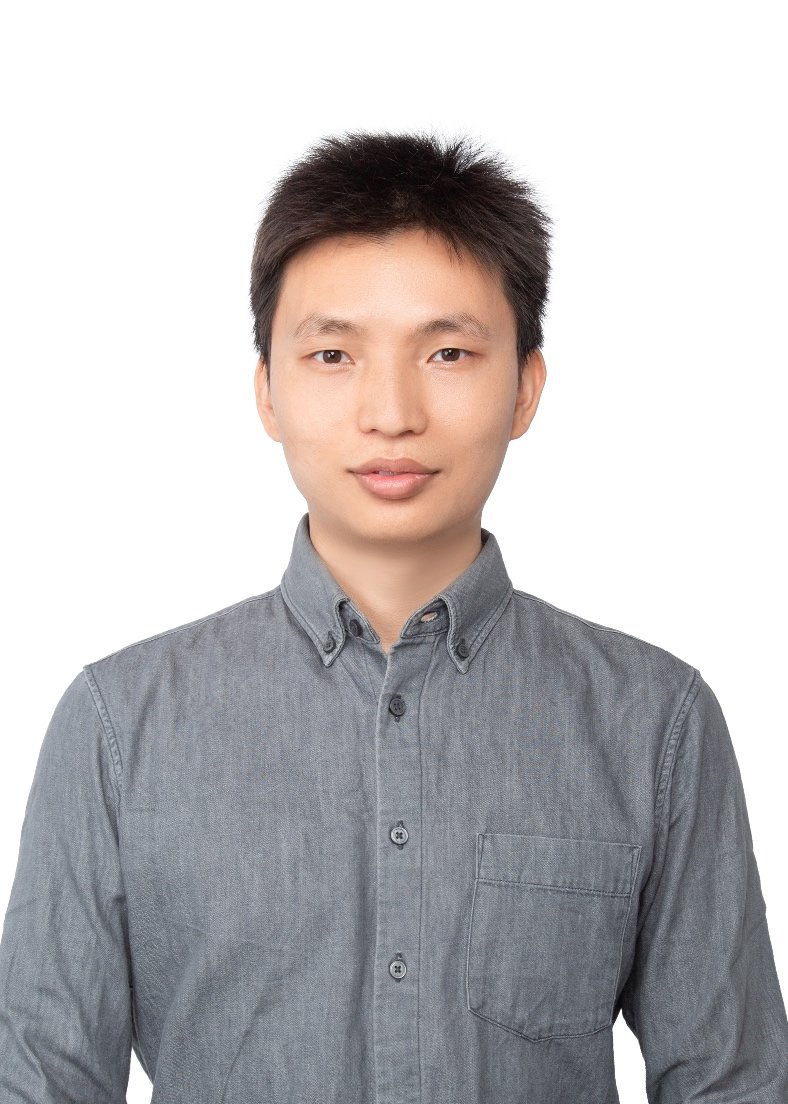}}]{Dingquan Li}(Member, IEEE) received the dual B.S.
degrees in electronic science and technology and
applied mathematics from Nankai University, Tianjin, China, in 2015, and the Ph.D. degree in applied mathematics from Peking University, Beijing, China, in 2021. He was a Postdoc and is currently an Associate Researcher with Pengcheng Laboratory, Shenzhen, China. His research interests include multimedia processing and machine learning, especially quality assessment, data compression, and perceptual optimization.
\end{IEEEbiography}

\begin{IEEEbiography}[{\includegraphics[width=1in,height=1.25in,clip,keepaspectratio]{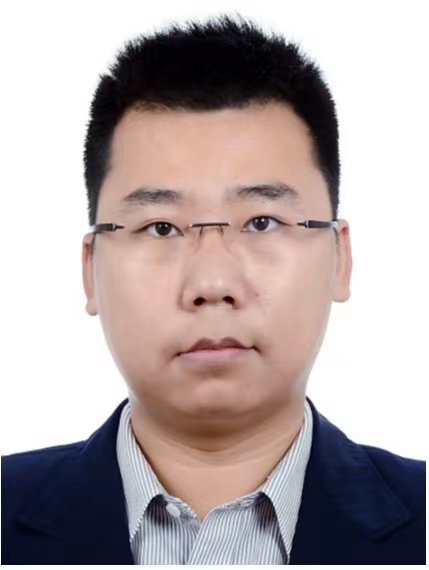}}]{Guangtao Zhai}(Fellow, IEEE) received the B.E. and M.E. degrees from Shandong University, Shandong, China, in 2001 and 2004, respectively, and the Ph.D. degree from Shanghai Jiao Tong University, Shanghai, China, in 2009, where he is currently a Research Professor with the Institute of Image Communication and Information Processing. From
2008 to 2009, he was a Visiting Student with the Department of Electrical and Computer Engineering, McMaster University, Hamilton, ON, Canada, where he was a Post-Doctoral Fellow from 2010 to 2012. From 2012 to 2013, he was a Humboldt Research Fellow with the Institute of Multimedia Communication and Signal Processing, Friedrich Alexander University of Erlangen-Nuremberg, Germany. He received the Award of National Excellent Ph.D. Thesis from the Ministry of Education of China in 2012. His research interests include multimedia signal processing and perceptual signal processing.
\end{IEEEbiography}

\begin{IEEEbiography}[{\includegraphics[width=1in,height=1.25in,clip,keepaspectratio]{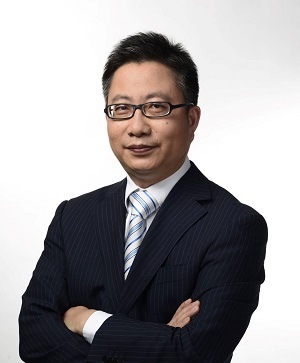}}]{Xiaokang Yang}(Fellow, IEEE)
received the B.S. degree from Xiamen University, Xiamen, China, in 1994, the M.S. degree from the Chinese Academy of Sciences, Shanghai, China, in 1997, and the Ph.D. degree from Shanghai Jiao Tong University, Shanghai, in 2000. From September 2000 to March 2002, he worked as a Research Fellow with the Centre for Signal Processing, Nanyang Technological University, Singapore. From April 2002 to October 2004, he was a Research Scientist with the
Institute for Infocomm Research (I2R), Singapore. From August 2007 to July 2008, he visited the Institute for Computer Science,
University of Freiburg, Germany, as an Alexander von Humboldt Research Fellow. He is currently a Distinguished Professor with the School of Electronic Information and Electrical Engineering, Shanghai Jiao Tong University. His current research interests include image processing and communication, computer vision, and machine learning. He is an Associate Editor of the IEEE TRANSACTIONS ON MULTIMEDIA.
\end{IEEEbiography}

\begin{IEEEbiography}[{\includegraphics[width=1in,height=1.25in,clip,keepaspectratio]{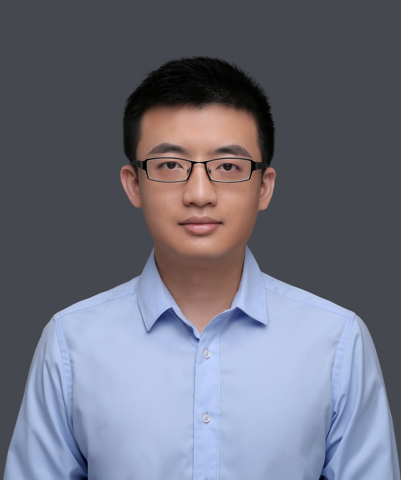}}]{Kede Ma} (Senior Member, IEEE) received the B.E. degree from the University of Science and Technology of China (USTC) in 2012, and the MASc. and Ph.D. degrees from the University of Waterloo in 2014 and 2017, respectively. From 2018 to 2019, he was a Research Associate with the Howard Hughes Medical Institute and New York University. He is currently an Associate Professor with the Department of Computer Science, City University of Hong Kong. His research interests include computational vision, computational photography, multimedia forensics and security, machine learning for multimedia signals, and multimedia data engineering. He currently serves on the Editorial Boards of IEEE Transactions on Image Processing, IEEE Transactions on Information Forensics and Security, and IEEE Signal Processing Letters.
\end{IEEEbiography}


\end{document}